\documentclass{article} % For LaTeX2e
\usepackage{iclr2027_conference,times}

\usepackage{amsmath,amsfonts,bm}

\def\eqref#1{equation~\ref{#1}}
\def\1{\bm{1}}

\DeclareMathAlphabet{\mathsfit}{\encodingdefault}{\sfdefault}{m}{sl}
\SetMathAlphabet{\mathsfit}{bold}{\encodingdefault}{\sfdefault}{bx}{n}

\usepackage{hyperref}
\usepackage{url}
\usepackage{enumitem}
\usepackage{amsmath,amssymb}
\usepackage{algorithm}
\usepackage{algpseudocode}
\usepackage{booktabs}
\usepackage{multirow}
\usepackage{natbib}
\usepackage{graphicx}
\usepackage{subcaption}
\usepackage{array}
\usepackage{tabularx}
\usepackage{makecell}
\usepackage{xcolor}
\usepackage{color}
\usepackage{comment}
\usepackage{nicefrac}

\usepackage{wrapfig}
\usepackage{caption}
\usepackage{booktabs}

\newcolumntype{Y}{>{\raggedright\arraybackslash}X}

\usepackage{wrapfig}  % wraptable：文字环绕表格
\usepackage{amssymb} % \checkmark
\usepackage{fontawesome5}
\usepackage{xcolor}
\usepackage{hyperref}

\definecolor{linkblue}{RGB}{0,102,204}

\hypersetup{
    colorlinks=true,
    linkcolor=black,
    citecolor=black,
    urlcolor=linkblue
}

\newcommand{\equalmark}{*}
\newcommand{\leadmark}{\ensuremath{\dagger}}
\newcommand{\corrmark}{\faEnvelope}

\definecolor{skyblue}{RGB}{135,206,235}
\definecolor{brickred}{RGB}{203,65,84}
\newcommand{\anlin}[1]{{\color{black}#1}}

\color{black}
\title{D$^2$-VLA: Dual-Memory Dual-Frequency Vision-Language-Action Model For Long Dynamic Manipulation}

\author{
\makebox[\textwidth][c]{%
{\bfseries
Zijian Ye\textsuperscript{1,\equalmark},
Chengqi Wei\textsuperscript{2,\equalmark},
Wei Huang\textsuperscript{1,\leadmark},
Anlin Zheng\textsuperscript{1,\leadmark},
Chunyu Zou\textsuperscript{1},
Liangyu Wu\textsuperscript{2}
}%
}
\\
\makebox[\textwidth][c]{%
{\bfseries
Zikang Zhao\textsuperscript{2},
Zhenjie Peng\textsuperscript{2},
Yushuo Yang\textsuperscript{2},
Shuman Zhao\textsuperscript{1},
Zhongrui Wang\textsuperscript{2,\corrmark},
Xiaojuan Qi\textsuperscript{1,\corrmark}
}%
}
\\[0.9em]
\makebox[\textwidth][c]{%
{\normalfont
\textsuperscript{1} The University of Hong Kong
\qquad
\textsuperscript{2} Southern University of Science and Technology
}%
}
\\[0.5em]
\makebox[\textwidth][c]{%
  {\fontsize{9pt}{11pt}\selectfont
    \faIcon{github}\hspace{0.35em}%
    \href{https://github.com/CVMI-Lab/DMDF-VLA}{%
      \textbf{\texttt{D}\textsuperscript{2}\texttt{-VLA}}%
    }%
    \hspace{1.5em}%
    \faIcon{rocket}\hspace{0.35em}%
    \href{https://zijianyy.github.io/Dual-Memory-Dual-Frequency-VLA-Webpage/}{%
      \textbf{\texttt{D}\textsuperscript{2}\texttt{-VLA}}%
    }%
  }%
}
}

\iclrfinalcopy 
\begin{document}

\maketitle

\lhead{}
% Author contribution notes at the bottom of the first page
% Author-role notes at the bottom of the first page
\begingroup
\renewcommand{\thefootnote}{}
\footnotetext{%
\textsuperscript{*} Equal contribution.
\ \textsuperscript{\ensuremath{\dagger}} Project lead.
\ \textsuperscript{\faEnvelope} Corresponding author:
\href{mailto:corresponding@example.ac.jp}
{\texttt{wangzr@sustech.edu.cn, xjqi@eee.hku.hk}}.
}
\addtocounter{footnote}{-1}
\endgroup

\begin{abstract}
Long-horizon manipulation requires robots to remember cues that are no longer in view while responding to moving objects. Yet vision-language-action (VLA) policies often rely on the latest observation, and refreshing their visual context typically requires another costly vision-language model (VLM) pass. We present D$^2$-VLA, which combines dual memory and dual-frequency control at the KV-cache interface of a pretrained VLA. D$^2$-VLA uses block-wise causal KV caching to encode observations incrementally and, guided by distinct temporal attention patterns, constructs separate historical KV read views for the VLM and action expert. Between periodic VLM updates, a gated adapter incorporates fresh visual features into the latest history-conditioned KV block, while a short fast-memory queue supports action replanning. We introduce DOMINO-Long, a ten-task benchmark requiring robots to use earlier visual cues when manipulating moving objects. D$^2$-VLA achieves complete-task success rates of 29.3\% on DOMINO, compared with 9.6\% for $\pi_{0.5}$ and 17.2\% for PUMA, and 60.0\% on DOMINO-Long, compared with 35.4\% and 20.6\%, respectively. It improves success rates on eight real-robot tasks and reaches 97.5\% on LIBERO-Long and 74.3\% on RoboTwin 2.0.

% We also introduce DOMINO-Long, a ten-task benchmark that requires robots to use earlier visual cues when interacting with moving objects. D$^2$-VLA achieves a \textbf{29.3\%} complete-task success rate on DOMINO, outperforming $\pi_{0.5}$ (9.6\%) by \textbf{19.7\%} and PUMA (17.2\%) by \textbf{12.1\%}. On DOMINO-Long, it reaches 60.0\%, improving over $\pi_{0.5}$ (35.4\%) by \textbf{24.6\%} and PUMA (20.6\%) by \textbf{39.4\%}. D$^2$-VLA also improves performance across 8 real-robot tasks and achieves \textbf{97.5\%} on LIBERO-Long and \textbf{74.3\%} on RoboTwin 2.0, respectively.

\end{abstract}

%The memory of VLA needs to be fast and Vision-language-action (VLA) models often rely on the latest observation, limiting their ability to retain earlier cues and track task progress in long-horizon manipulation. Simply extending the observation history increases computational overhead, while incorporating fresh visual information typically requires a full vision-language model (VLM) pass. These limitations make it difficult to combine long-term visual memory with timely responses to dynamic environments. In this work, we introduce D$^2$-VLA, a framework combining dual memory and dual-frequency control for long-horizon dynamic manipulation. Dual-memory strategy provides component-specific historical context, preserving broader temporal information for the VLM while emphasizing recent observations for the action expert. Dual-frequency control decouples slow VLM updates from fast action replanning through a lightweight KV adapter that incorporates current visual evidence between VLM updates, supported by a short fast-memory queue. This design combines historical context with fresh visual feedback without requiring a full VLM pass at every replanning step. We further introduce Domino-Long, a benchmark targeting both temporal memory and dynamic responsiveness. D$^2$-VLA achieves success rates of 29.3\% on DOMINO and 60.0\% on DOMINO-Long, exceeding $\pi_{0.5}$ by 19.7 and 24.6 percentage points, respectively, and outperforms both $\pi_{0.5}$ and MemoryVLA on all four long-horizon dynamic real-world tasks.

\section{Introduction}
\label{sec:introduction}

\begin{figure}[htbp]
    \centering
    \includegraphics[width=0.99\linewidth]{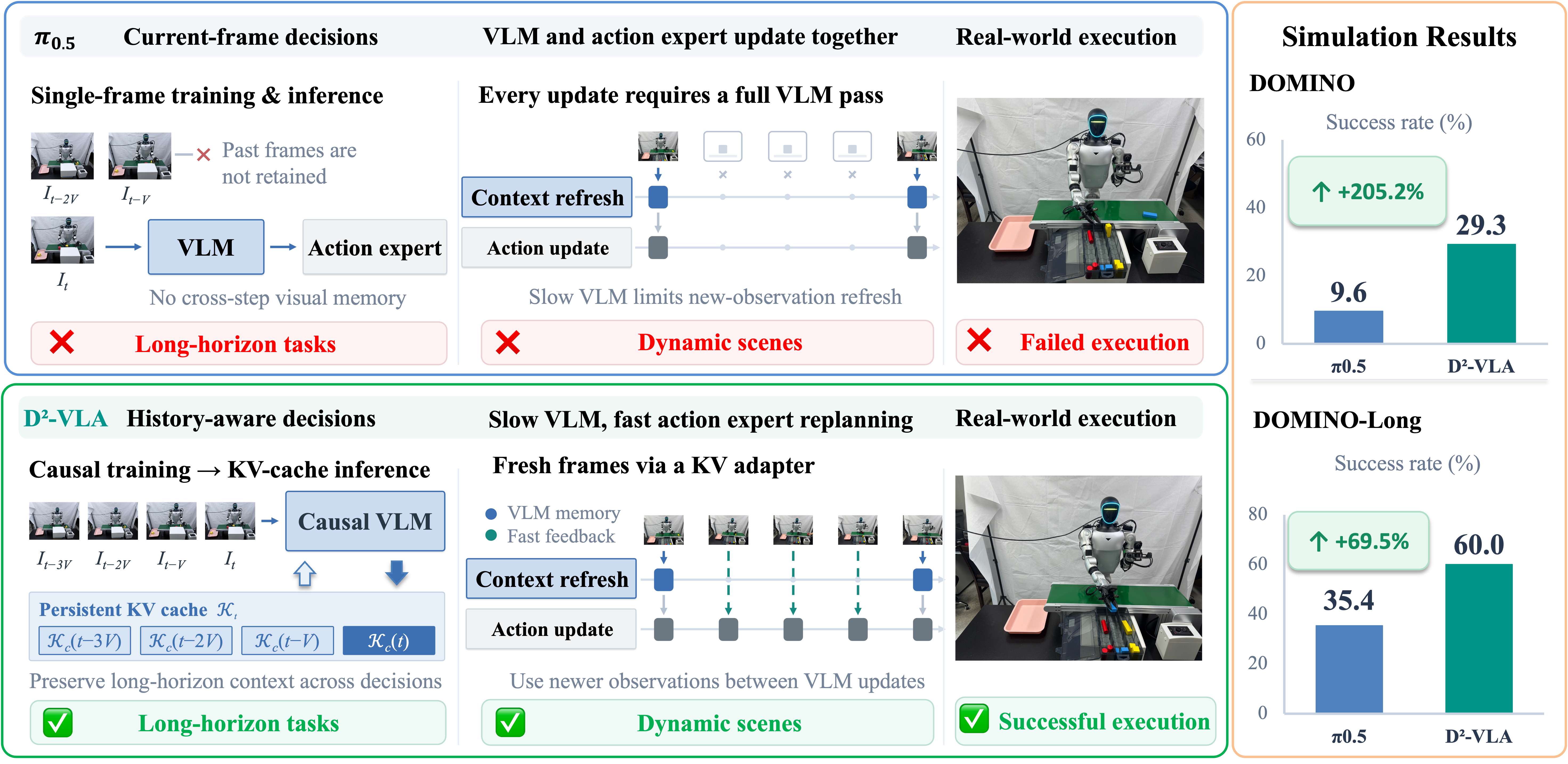}
    \vspace{-.5pc}
    \caption{
    \textbf{Persistent memory and responsive control with D$^2$-VLA.}
    Unlike single-frame $\pi_{0.5}$ (top), D$^2$-VLA (bottom) retains temporal context through causal KV caching and incorporates fresh observations through a KV adapter for fast replanning between slow VLM updates.
    Real-world examples and benchmark success rates illustrate improvements in long-horizon dynamic manipulation.
    }
    \label{fig:intro}
    \vspace{-3.pc}
\end{figure}

Vision--language--action (VLA) models leverage pretrained vision--language models (VLMs) to map visual observations and language instructions to robot actions, enabling broad generalization across tasks, objects, and environments \citep{intelligence2025pi05visionlanguageactionmodelopenworld,kim2024openvlaopensourcevisionlanguageactionmodel}. Modern flow-based VLAs, such as  $\pi_{0.5}$, naturally separate computation into a large VLM for semantic understanding and an action expert for continuous control. However, their inference loop is typically both \emph{memoryless} and \emph{frequency-coupled}: each action prediction relies primarily on the latest observation, and refreshing the action expert's visual context requires another full VLM pass, as shown
 in Fig.~\ref{fig:intro}.

% Paragraph 2: Coupled challenges in long-horizon and dynamic manipulation.
\begin{comment}
Long-horizon manipulation becomes memory-dependent when task-relevant information is no longer available in the current observation: a robot may need to remember an earlier cue, distinguish completed from pending subtasks, or recover an object's identity after occlusion.
Under these requirements, visually similar states can require different actions depending on the preceding interaction.
Dynamic environments impose a second requirement: the policy must incorporate new visual evidence quickly enough to respond to object motion changes.
These demands often arise together, as when a robot must retain an earlier instruction cue while reaching for a moving target.
A longer observation history can reduce temporal ambiguity, but repeatedly encoding that history increases computation.
Conversely, executing longer action chunks amortizes inference cost while delaying the next opportunity for visual correction~\citep{black2025realtimeexecutionactionchunking}.
In the baseline inference loop, refreshing the action expert's visual conditioning also requires a new VLM pass.
Effective control therefore requires both access to relevant past observations and timely incorporation of the present. 
\end{comment}

However, real-world manipulation poses two coupled temporal challenges: \textit{long-horizon memory}, because task cues, object identities, and progress may disappear from the current view; and \textit{fast responsiveness}, because objects are not always static. For instance, grasping a rolling bottle and returning it to its original position requires both. Simply retaining longer VLM context histories increases attention costs, whereas frequent VLM refreshes are expensive. Therefore, effective control needs to know \textit{what history} the VLM and action expert access, and \textit{how frequently} each is updated.

Prior work has largely studied these two dimensions separately. Memory-oriented methods aggregate recent frames \citep{li2025cronusvlaefficientrobustmanipulation}, retrieve perceptual and semantic history \citep{shi2026memoryvlaperceptualcognitivememoryvisionlanguageaction}, propagate recurrent states \citep{li2026rememvlaempoweringvisionlanguageactionmodel}, or combine short-term visual context with long-term semantic summaries or selected keyframes \citep{torne2026memmultiscaleembodiedmemory,sridhar2025memerscalingmemoryrobot}. KV-based approaches further retain or retrieve historical keys and values \citep{sun2026tempofitplugandplaylayerwisetemporal}, reducing computational cost, but they do not explicitly account for the potentially different histories required by the VLM (for semantic reasoning) and the action expert (for action generation). In parallel, real-time and hierarchical approaches improve responsiveness through execution overlap \citep{black2025realtimeexecutionactionchunking,xie2026dynamicvlavisionlanguageactionmodeldynamic} or slow--fast computation \citep{zhang2025hirtenhancingroboticcontrol,li2026favlaforceadaptivefastslowvla}. Their fast controllers, however, are typically conditioned on the latest slow representation. AHA-WAM maintains rolling KV memory in a low-frequency video-DiT world planner and adapts its context to current observations for a high-frequency action DiT \citep{cai2026ahawamasynchronoushorizonadaptiveworldactionmodeling}. However, a low-level WAM policy typically does not need long video memory for higher-level semantic reasoning. These distinctions lead to our central question: \emph{can memory access and update frequency be co-designed at the native KV interface of a pretrained VLA?} 

To investigate this question, we extend the single-observation $\pi_{0.5}$ policy with block-causal temporal modeling and observe a pronounced temporal asymmetry: the VLM relies on broader history for semantic reasoning, whereas the action expert attends mostly to token subsets from recent observations that contain motion information, as shown in Fig.~\ref{fig:pattern}. These findings motivate decoupling KV-cache selection and refresh rates for the VLM and action expert.

We therefore introduce \textbf{D}ual-Memory \textbf{D}ual-Frequency VLA (D$^2$-VLA) (Fig.~\ref{fig:intro}).
Block-causal training equips a single-frame VLA with temporal memory, enabling inference to encode only the latest observation while reusing historical KV states.
\textit{Dual-Memory} uses consumer-specific online attention statistics to select historical KV for the VLM and action expert, retaining relevant context while reducing history-read and attention costs.
\textit{Dual-Frequency} combines slow VLM refreshes with fast action replanning: between refreshes, a lightweight gated adapter refines the latest history-conditioned KV block using fresh visual features, without rerunning the VLM.
A bounded fast-memory queue retains recent refined blocks to condition the action expert on short-range dynamics.

We further introduce \textbf{DOMINO-Long}, a benchmark whose tasks require robots to retain task information, such as earlier spatial configurations, presentation orders, or color cues, over long time scales while simultaneously manipulating dynamic objects which are constantly moving. Across DOMINO~\citep{fang2026generalizableroboticmanipulationdynamic}, DOMINO-Long, static long-horizon benchmarks, and real-robot experiments, D$^2$-VLA consistently improves over single-frame and memory-augmented baselines. It achieves a 29.3\% complete-task success rate and a 40.6 manipulation score on DOMINO, reaches 97.5\% on LIBERO-Long~\citep{liu2023liberobenchmarkingknowledgetransfer} and 74.3\% on RoboTwin~2.0~\citep{robotwin2}, and outperforms $\pi_{0.5}$ and MemoryVLA on all four reported dynamic real-robot tasks.
\begin{comment}
We introduce \emph{Dynamic-Long}, a dataset and evaluation suite targeting the joint demands of temporal memory and dynamic manipulation, and complement it with real-robot experiments.
\textbf{[Evaluation results to be added: Dynamic-Long success rates and comparisons; measured inference latency or throughput; ablations of KV selection and dual-frequency control; real-robot outcomes.]}
\end{comment}

Our main contributions are:
\begin{itemize}
[
    labelindent=0pt,
    leftmargin=*,
    labelsep=0.5em,
    itemsep=2pt,
    align=left
]
\item We uncover a previously underexplored temporal asymmetry: the VLM and action expert require distinct historical contexts and memory update rates. Motivated by this finding, we introduce {D$^2$-VLA}, a unified dual-memory, dual-frequency framework.
\item We design a \textit{Dual-Memory} method that enables the VLM and action expert to selectively retrieve KV from cache based on attention statistics, preserving relevant history while reducing redundant context computation.
\item We develop a \textit{Dual-Frequency} mechanism, which combines periodic VLM updates with lightweight gated visual adaptation, fast-memory buffering, and grouped fine-tuning, enabling responsive action replanning without rerunning the full VLM.
\item We introduce \textbf{DOMINO-Long}, a benchmark demanding both long-term semantic reasoning and dynamic object interaction, and demonstrate consistent gains across comprehensive simulation and real-robot evaluations.
\end{itemize}

\begin{comment}
Our main contributions are:
\begin{itemize}
    \item \textbf{Causal temporal adaptation with dual-memory selection.}
    We equip a single-frame VLA with observation-level autoregressive memory through block-causal training and incremental KV inference.
    Guided by the different temporal attention patterns of the VLM and action expert, we design separate historical KV selection rules that reduce the context each component reads while retaining access to task-relevant history.

    \item \textbf{Dual-frequency control with current visual conditioning.}
    We decouple VLM and action-expert refresh rates through a gated KV adapter, canonical slow memory, and a short fast-history queue.
    A grouped fine-tuning strategy with randomized temporal offsets trains the adapter and action expert to use recent observations together with slowly updated context.

    \item \textbf{Evaluation of long-horizon dynamic manipulation.}
    We construct Dynamic-Long to study tasks that require both memory of earlier observations and responses to evolving scenes, and assess the approach through comparative evaluation, component ablations, and real-robot experiments.
    \textbf{[Benchmark specifications and experimental findings to be completed.]}
\end{itemize}
\end{comment}
\section{Related Work}
\label{sec:related_work}

\paragraph{Long-horizon manipulation and memory-augmented policies.}
History-dependent manipulation requires information absent from the current observation. RoboMME evaluates temporal, spatial, object, and procedural memory~\citep{dai2026robommebenchmarkingunderstandingmemory}. RoboFlamingo models history with a sequential policy head~\citep{li2024visionlanguagefoundationmodelseffective}, while CronusVLA extends single-frame VLAs through multi-frame post-training~\citep{li2025cronusvlaefficientrobustmanipulation}. MemoryVLA retrieves perceptual and cognitive memories for action generation~\citep{shi2026memoryvlaperceptualcognitivememoryvisionlanguageaction}; MEM combines short-term video memory with long-term textual memory~\citep{torne2026memmultiscaleembodiedmemory}. MemER selects historical keyframes to guide a low-level policy through language instructions~\citep{sridhar2025memerscalingmemoryrobot}. We instead study consumer-specific historical KV selection, preserving complete stored blocks while tailoring their reads.

\paragraph{Dynamic manipulation and real-time control.}
Dynamic manipulation requires both timely sensory updates and effective
task understanding. Recent works address this challenge through
different forms of temporal adaptation. RTC reduces execution latency
by overlapping action generation and execution without retraining
~\citep{black2025realtimeexecutionactionchunking}, while DynamicVLA
introduces a dynamic manipulation benchmark and a compact VLA with
streaming inference~\citep{xie2026dynamicvlavisionlanguageactionmodeldynamic}.
HiRT separates fast visual control from slowly updated VLM context
through hierarchical computation~\citep{zhang2025hirtenhancingroboticcontrol},
and FAVLA reuses VLM KV states while incorporating high-frequency force
feedback into action generation~\citep{li2026favlaforceadaptivefastslowvla}.
These studies highlight the importance of adaptive update schedules
and efficient sensory conditioning for dynamic environments.
Our dual-frequency control further explores how current visual
observations can efficiently refresh a history-conditioned VLA through
a lightweight KV refinement interface.

\section{Method}
\label{sec:method}

\begin{figure}[htbp]
    \centering
    \includegraphics[width=0.99\linewidth]{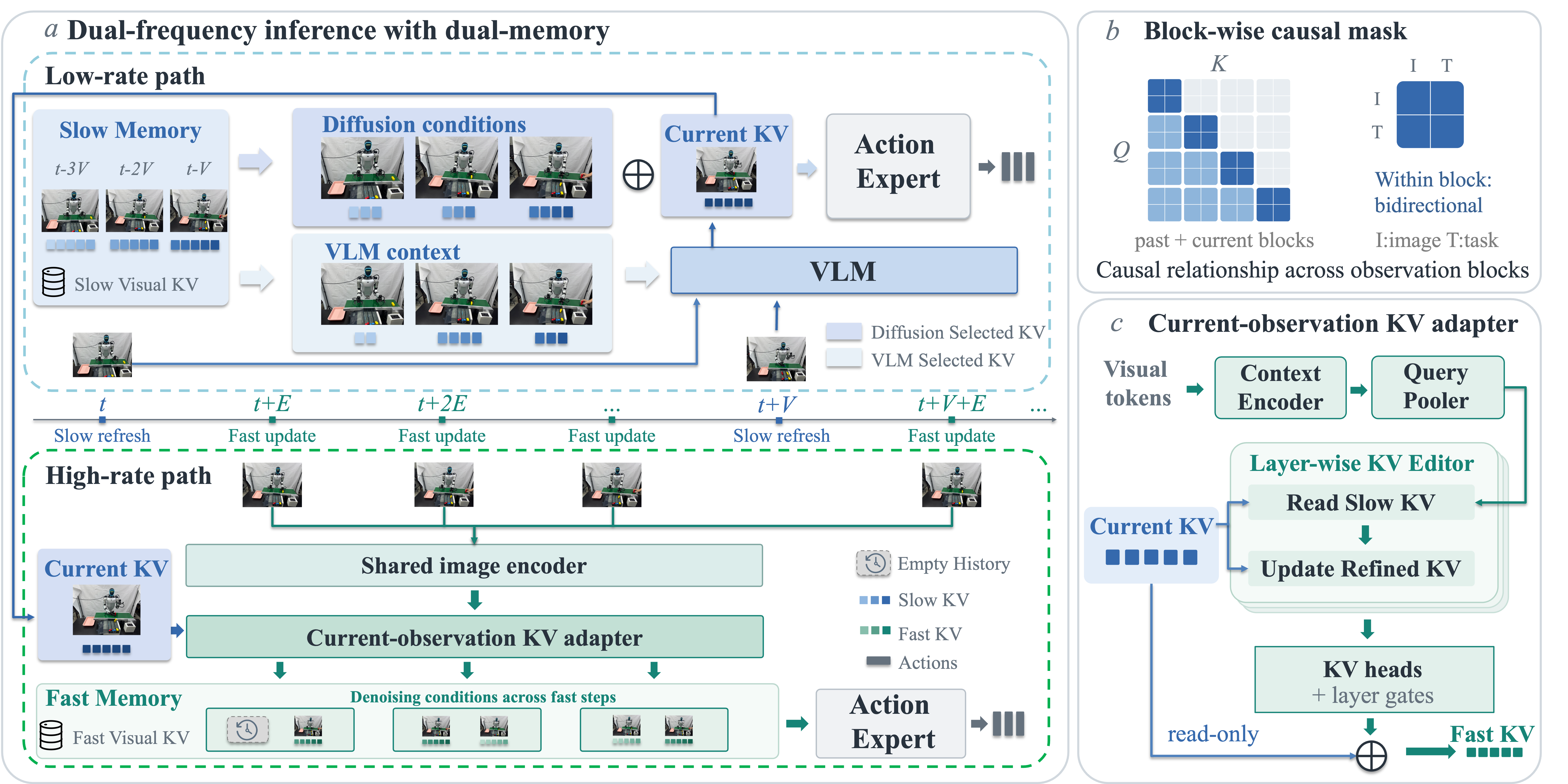}
    \vspace{-.5pc}
    \caption{
    \textbf{Overview of D$^{2}$-VLA.}
    \textbf{(a)} Dual memory provides component-specific KV reads; dual-frequency control combines slow VLM refreshes with fast action replanning.
    \textbf{(b)} Block-causal attention allows bidirectional reads within each observation block.
    \textbf{(c)} The adapter refines the latest slow KV block with current visual features.
    }
    \label{fig:method_overview}
    \vspace{-1.5pc}
\end{figure}

We first formulate the task, then introduce block-wise causal KV reuse, component-specific historical read views, and dual-frequency visual conditioning. Implementation details and numerical configurations are provided in Appendix~\ref{app:implementation}.

\subsection{Preliminaries}
\label{sec:preliminaries}

\paragraph{Problem formulation.}

{Generally, at each environment step $t$, the robot receives multi-view visual observations $I_t$, a language instruction $\ell$, and its proprioceptive state $s_t$.
Let $C$ denote the number of camera views, $H$ the predicted action horizon, and $E$ the executed chunk size.
The observation set and predicted action chunk are
\begin{equation}
 I_t=\{I_t^{(c)}\}_{c=1}^{C},\qquad
 A_t=(a_t,\ldots,a_{t+H-1}),\qquad 1\leq E\leq H.
 \label{eq:observation_action}
\end{equation}
Here $a_t$ denotes an action at environment step $t$.
The policy executes the first $E$ actions and subsequently replans from the updated observation.
We denote the vision--language model by $\Phi$ and the flow-matching action expert with parameters $\theta$ by $v_\theta$; $\pi_\theta$ denotes the action distribution induced by flow integration.
In D$^2$-VLA, the current proprioceptive state is projected into a dedicated token and supplied directly to the action expert, rather than included in the VLM observation blocks or their cached history. For notational simplicity, we leave this state input implicit in the policy equations below.
In our single-frame configuration initialized from $\pi_{0.5}$,
action generation is conditioned on the current observation without historical context:
\begin{equation}
A_t^{\mathrm{single}}
\sim
\pi_\theta\!\left(
\cdot \mid \Phi(I_t, \ell, s_t)
\right).
\label{eq:single_frame}
\end{equation}
Here, the VLM encodes the current images and instruction into task-conditioned representations, which are subsequently consumed by the action expert to jointly generate the entire action chunk.
}

\paragraph{Limited temporal and high-frequency visual context.}

{Single-frame policies such as $\pi_{0.5}$~\citep{intelligence2025pi05visionlanguageactionmodelopenworld} lack historical context for handling occlusions, tracking task progress, and making long-horizon decisions. Refreshing their visual conditioning requires a full VLM pass at each replanning step. Re-encoding history adds computational cost, while reusing stale features can reduce action accuracy. This trade-off between temporal context, fresh visual feedback, and inference cost motivates D$^2$-VLA.}

\subsection{Causal Temporal Modeling}
\label{sec:causal_modeling}

\paragraph{Task-conditioned observation blocks.}
Within an episode, we sample $N$ observations at interval $\Delta$, including the current observation:
\begin{equation}
\mathcal{T}_t =
\left[
I_{t-(N-1)\Delta},
\ldots,
I_{t-\Delta},
I_t
\right].
\end{equation}
The image and text encoders, $E_{\mathrm{img}}$ and $E_{\mathrm{text}}$, form Image--Task blocks $B_j$ with the instruction $\ell$ repeated at each sampled time $j$:
\begin{equation}
B_j=\left[E_{\mathrm{img}}(I_j);E_{\mathrm{text}}(\ell)\right],\qquad
\mathcal{B}_{\mathcal{T}}=\left[B_{t-(N-1)\Delta};\ldots;B_t\right].
\label{equ:encoding_obs}
\end{equation}
Here $E_{\mathrm{img}}(I_j)$ concatenates tokens from all camera views, semicolons denote token concatenation, and $\mathcal{B}_{\mathcal{T}}$ is the resulting multimodal sequence.

\paragraph{Block-wise causal attention.}
The VLM uses bidirectional attention within each block and causal attention across blocks.
Its observation KV and the separately projected current state condition the action expert; the full attention mask is given in Appendix~\ref{app:causal_details}.
At inference, retained history $\mathcal{K}_{<t}$ is reused without re-encoding, and the current block's complete visual and instruction KV, $\mathcal{K}_c$, is appended to form $\mathcal{K}_t$:
\begin{equation}
\mathcal{K}_c=\Phi_{\mathrm{KV}}\!\left(I_t,\ell;\mathcal{K}_{<t}\right),\qquad
\mathcal{K}_t=\mathcal{K}_{<t}\oplus \mathcal{K}_c,\qquad
A_t^{\mathrm{hist}}\sim\pi_\theta\!\left(\cdot\mid\mathcal{K}_t\right).
\label{equ:cache_queue}
\end{equation}
Here $\Phi_{\mathrm{KV}}$ returns layer-wise KV and $\oplus$ denotes temporal concatenation; $c$ identifies the current block, not a camera, and $\mathcal{K}_c(j)$ specifies its observation time.
Each update uses up to the latest $N-1$ historical blocks, using all available history during warm-up.
We call this policy the \textit{KV-Cache} prototype.

\begin{figure}[htbp]
    \centering
    \includegraphics[width=0.99\linewidth]{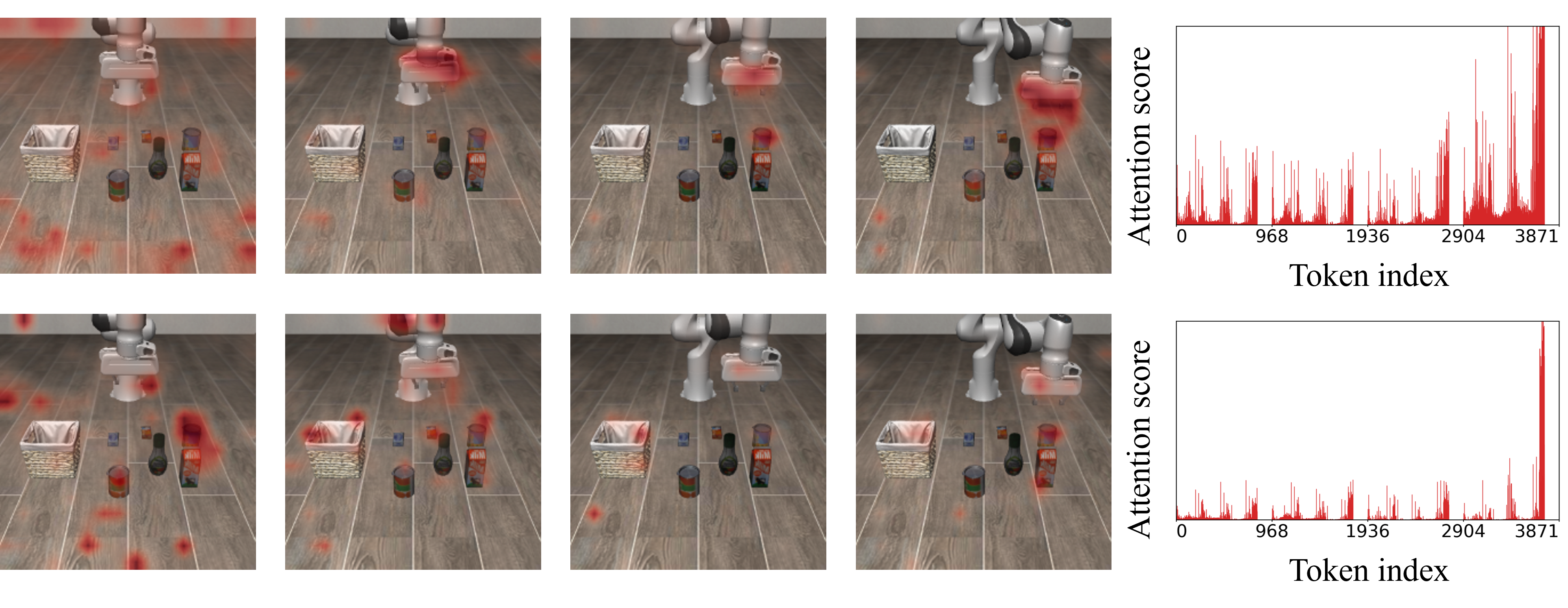}
    \vspace{-1.2pc}
    \caption{
    \textbf{Attention asymmetry and token concentration.}
    Visual attention maps (left) and token-wise attention scores (right) for the VLM (top) and action expert (bottom).
    }
    \vspace{-2pc}
    \label{fig:pattern}
\end{figure}

\subsection{Dual-Memory: Asymmetric History Selection}
\label{sec:dual_memory_selection}

The \textit{KV-Cache} prototype improves success over single-observation $\pi_{0.5}$ on LIBERO-Long and RoboTwin~2.0, but increases latency and memory (Tab.~\ref{tab:kv_cache_latency}).
Therefore, we analyze its attention patterns to identify opportunities for selective history access.

\paragraph{Attention asymmetry and token concentration.}
Aggregating historical attention by temporal distance reveals different temporal preferences for the VLM and action expert (\textit{attention asymmetry}).
Within observations, both modules concentrate attention on a small subset of tokens (\textit{token concentration}; Fig.~\ref{fig:pattern}).
These findings motivate independent historical KV reads for the two modules; additional visualizations are provided in Appendix~\ref{app:attention_analysis}.

\paragraph{Asymmetric historical read views.}
The VLM and action expert use $\operatorname{Read}_{\mathrm{VLM}}$ and $\operatorname{Read}_{\mathrm{act}}$, respectively, to select historical visual tokens from complete stored blocks $\mathcal{K}_{<t}$ using independent online attention statistics, while retaining instruction tokens.
Selection is inference-only; training uses uncompressed KV.
The VLM reads its selected history to encode the current observation, while the expert combines its own selection with the complete current block:
\begin{equation}
 \mathcal{K}_c=\Phi_{\mathrm{KV}}\!\left(I_t,\ell;\operatorname{Read}_{\mathrm{VLM}}(\mathcal{K}_{<t})\right),\qquad
 \widehat{\mathcal{K}}_t^{\,\mathrm{act}}=\operatorname{Read}_{\mathrm{act}}(\mathcal{K}_{<t})\oplus \mathcal{K}_c.
 \label{eq:memory_views}
\end{equation}
Here $\widehat{\mathcal{K}}_t^{\,\mathrm{act}}$ is the expert's observation-KV condition.
Selection changes only temporary reads, leaving stored blocks complete and the current block uncompressed.
The condition remains fixed during each flow solve, whose attention statistics update subsequent reads.
Online scoring and token selection are detailed in Appendix~\ref{app:selection_details}.

\subsection{Dual-Frequency: Multi-Rate Visual Conditioning}
\label{sec:dual_frequency}

\anlin{
D$^2$-VLA updates VLM every $V$ environment steps and replans every $E$ steps, with $V$ an integer multiple of $E$.
A dynamic feature adapter supplies fresh visual conditioning between VLM refreshes.

\paragraph{Dual-memory mechanism.}
The queues $Q_{\mathrm{slow}}$ and $Q_{\mathrm{fast}}$ store complete VLM-generated and adapter-refined KV blocks with capacities $N_{\mathrm{slow}}$ and $N_{\mathrm{fast}}$.
Read operators act on their layer-wise temporal concatenations; for $Q_{\mathrm{slow}}$, this is $\mathcal{K}_t$ at slow-refresh steps.
VLM prefill, slow denoising, and fast denoising maintain independent selection statistics.
Before insertion, each queue keeps the most recent historical blocks up to one less than its capacity, denoted by $Q_{\mathrm{slow}}^{<t}$ or $Q_{\mathrm{fast}}^{<t}$.
A slow refresh generates and stores the complete current block:
\begin{equation}
\mathcal{K}_c
=
\Phi_{\mathrm{KV}}\!\left(
I_t,
\ell;
\operatorname{Read}_{\mathrm{VLM}}(Q_{\mathrm{slow}}^{<t})
\right),
\qquad
Q_{\mathrm{slow}}
=
Q_{\mathrm{slow}}^{<t}
\oplus
\mathcal{K}_c,
\label{eq:slow_anchor}
\end{equation}

\paragraph{Dynamic Feature Adapter.}
Let $t_s$ denote the latest slow-update time; within its cycle, we abbreviate the complete anchor $\mathcal{K}_c(t_s)$ from Eq.~\ref{eq:slow_anchor} as $\mathcal{K}_c$.
With steps counted from episode start, the normalized elapsed time $\delta_t$ and configured history interval $\Delta$ are
\begin{equation}
\Delta=V,\qquad {\delta}_{t}=\nicefrac{(t-\lfloor\nicefrac{t}{V}\rfloor\cdot{V})}{V}.
\label{eq:elapsed_phase}
\end{equation}
The encoder $\operatorname{Enc}_{\psi}$ combines current image features with the temporal offset, and the learned-query pooler $\operatorname{Pool}_{\psi}$ produces compact queries $\mathcal{Z}_t$.
These queries read the anchor to produce a complete refined KV block $\widetilde{\mathcal{Z}}_t$, which is appended to fast memory:
\begin{equation}
\mathcal{Z}_t=\operatorname{Pool}_{\psi}\!\left(\operatorname{Enc}_{\psi}(E_{\mathrm{img}}(I_t),\delta_t)\right),\;
\widetilde{\mathcal{Z}}_t=\mathcal{K}_c+G_{\psi}\odot\operatorname{F}_{\psi}\!\left(\mathcal{Z}_t,\mathcal{K}_c\right),\;
Q_{\mathrm{fast}}^{t}=Q_{\text{fast}}^{<t}\oplus\widetilde{\mathcal{Z}}_t.
\label{eq:fast_feature_update}
\end{equation}
Here $\psi$ denotes adapter parameters, $\operatorname{F}_{\psi}$ predicts full-block K/V residuals, $G_{\psi}$ contains layer-wise K/V gates, and $\odot$ denotes elementwise multiplication.
Each refinement preserves the KV layout and uses the unchanged complete slow anchor, not a compressed view or previous refinement.
Task information enters through the anchor, without a separate text input (Appendix~\ref{app:dual_details}).

\paragraph{Dual-frequency Mechanism.}
Every $V$ steps, Eq.~\ref{eq:slow_anchor} refreshes slow memory and resets the fast queue and its scores.
Intermediate replanning uses Eq.~\ref{eq:fast_feature_update} without rerunning the VLM or modifying slow memory.
At each $E$-step replanning call, the shared expert uses
\begin{equation}
\widehat{\mathcal{K}}_t^{\,\mathrm{act}}
=
\begin{cases}
\operatorname{Read}_{\mathrm{act}}
\!\left(Q_{\mathrm{slow}}^{<t}\right)
\oplus \mathcal{K}_c,
& t \bmod V = 0, \\[2mm]
\operatorname{Read}_{\mathrm{act}}
\!\left(Q_{\mathrm{fast}}^{<t}\right)
\oplus \widetilde{\mathcal{Z}}_t,
& t \bmod V \ne 0,
\end{cases}
\qquad
A_t \sim \pi_{\theta}\!\left(
\cdot \mid \widehat{\mathcal{K}}_t^{\,\mathrm{act}}
\right).
\label{eq:dual_rate_condition}
\end{equation}
Current state is supplied separately (Section~\ref{sec:preliminaries}).
The two queues are never concatenated into one expert condition; fast calls inherit long-range context through the history-conditioned slow anchor.
}

\section{DOMINO-Long Dataset}

\anlin{Existing DOMINO tasks primarily target short-horizon manipulation, where actions are executed within a limited interaction window and therefore place relatively weak demands on long-term memory and extended decision-making. To address this limitation, we introduce \textbf{DOMINO-Long}, a benchmark designed to evaluate whether VLA models can retain task-relevant information over long temporal intervals and use it for subsequent decisions under dynamic interaction.

DOMINO-Long contains 10 multi-stage manipulation tasks with 50 demonstrations per task, yielding 500 demonstrations and 227,208 observation--action records in total. Each demonstration spans 201--911 interaction steps, with an average length of 454.4 steps. The tasks explicitly separate information acquisition from action execution: cues such as previous object locations, presentation order and color instructions are observed early in an episode but are required only at later stages. Representative tasks include restoring an earlier object configuration, executing actions according to a previously observed sequence, and selecting a moving object based on a cue that is no longer visible. Several tasks additionally involve moving objects, requiring the policy to combine retained historical information with current visual feedback. In this way, DOMINO-Long evaluates long-horizon context retention and dynamic responsiveness. Further details are provided in Appendix~\ref{appendix:domino_long_dataset}.}

\section{Experiments}
\label{sec:experiments}

Our evaluation examines whether D$^2$-VLA can retain task context over long horizons while incorporating fresh visual feedback during dynamic manipulation.
We consider DOMINO for dynamic interaction, LIBERO-Long for multi-stage manipulation, and our DOMINO-Long benchmark for long-horizon dynamic tasks.
Eight real-world tasks provide complementary physical evaluation, while RoboTwin~2.0 serves as a supplementary manipulation benchmark.

\subsection{Experimental Setup}
\label{sec:exp-setup}

\paragraph{Model configuration.}
D$^2$-VLA adds a full-block visual KV adapter to $\pi_{0.5}$ Base, retaining its PaliGemma-3B VLM and shared flow-matching expert.
For DOMINO, DOMINO-Long, and real-world tasks, the slow and fast memories hold up to four and two complete KV blocks, respectively, including the current block.

\paragraph{Model training.}
\anlin{Each training sample is constructed within a single episode and contains a slow anchor, up to three historical observations, and three action branches: one at the anchor time and two at later steps. Each branch has its own images, robot state, and $H$-step action target. The anchor branch is processed with one block-causal VLM forward pass. The later branches reuse adapter-refined KV features from the shared anchor and, when available, from the previous refined block. We train all modules end-to-end using only the flow-matching loss, while masking unavailable history and action targets beyond the episode boundary. No auxiliary adapter loss or training-time KV selection is used. We optimize with AdamW using 40 anchor groups per batch for 75K updates on DOMINO and 50K updates on DOMINO-Long and real-world tasks. Additional training details are provided in Appendix~\ref{app:training_details} and Tab.~\ref{tab:training_settings}.}

\paragraph{Model inference.}
On DOMINO, DOMINO-Long, and real-world tasks, we fix $H=E=25$ and $V=75$, with adapter updates between VLM refreshes.
Inference uses EMA weights and ten Euler steps with a fixed KV condition per call.
For the main results, VLM prefill, slow denoising, and fast denoising each retain 70\% of historical visual tokens using independent scores, preserving instruction tokens, the current denoising block, and full stored KV.
LIBERO-Long and RoboTwin~2.0 disable the high-rate path while retaining full-KV training and online selection (Section~\ref{sec:exp-simulation}).

\anlin{
\paragraph{Evaluation Scenarios and Benchmarks.}
We evaluate D$^2$-VLA in dynamic simulation, real-world manipulation, and static simulation to test responsiveness, temporal memory, and execution.
\begin{itemize}
% [
%     leftmargin=0pt,
%     labelindent=0em,
%     labelsep=0.5em,
%     itemsep=3pt
% ]
[
    labelindent=0pt,
    leftmargin=*,
    labelsep=0.5em,
    itemsep=2pt,
    align=left
]

\item \textbf{Dynamic simulation.} DOMINO~\cite{fang2026generalizableroboticmanipulationdynamic} comprises 35 dynamic tasks built on RoboTwin~2.0~\cite{robotwin2} and SAPIEN~\cite{sapien}. We use only the Aloha-AgileX embodiment, training on clean setting and evaluating on clean L1 to test responses to moving objects. DOMINO-Long (Section~4) adds 10 tasks requiring earlier observations to guide later actions, with 50 demonstrations per task used for training.

\item \textbf{Real-world manipulation.} Eight long-horizon tasks on Unitree G1D comprise 4 static and 4 dynamic tasks. They test ordered execution, object assignments, and spatial memory; dynamic tasks additionally require remembered cues and fresh visual feedback during manipulation.

\item \textbf{Static simulation.} LIBERO-Long tests instruction grounding and multi-stage execution across 10 tasks, with 50 demonstrations per task. RoboTwin~2.0 tests bimanual manipulation across 50 tasks; we train only on each task's Aloha clean dataset and evaluate all 50 tasks in the clean setting.

\end{itemize}

Detailed evaluation protocols and task definitions are provided in Appendices~\ref{app:implementation},~\ref{app:simulation_protocols} and~\ref{sec:real_world_tasks}.
}

\subsection{Evaluation Metrics}
\label{sec:eval-metrics}

% \paragraph{Task performance.}
% Complete-task success rate (SR), our primary simulation and real-world metric, is the percentage of episodes satisfying all task conditions, averaged across tasks.
% Long-horizon success requires the full sequence, including required identities, ordering, and final configuration; intermediate manipulation or cue recognition alone is insufficient.

% \paragraph{Manipulation quality.}
% DOMINO's Manipulation Score (MS) measures quality beyond binary success, combining spatial progress toward the target (route completion) with penalties for leaving the safe workspace or field of view and colliding with clutter, following its original protocol.
\paragraph{Complete-task success rate (SR).}
It measures whether the policy completes the task. An episode is counted as successful only when all task requirements are satisfied. For long-horizon tasks, this includes completing the full action sequence with the correct object identities, ordering, and final configuration; partial execution or correct cue recognition alone does not count as success.

\paragraph{Manipulation Score (MS).}
DOMINO's Manipulation Score (MS) measures manipulation quality beyond binary task completion. MS reflects how effectively the robot progresses toward the target while penalizing unsafe or undesirable behaviors, such as leaving the valid workspace or field of view and colliding with surrounding clutter, following the original DOMINO protocol.

\begin{table*}[t]
    \centering
    % Shared settings keep the two tables visually consistent.
    \footnotesize
    \setlength{\tabcolsep}{3pt}
    \renewcommand{\arraystretch}{1.15}

    % ==================== Left table ====================
    \begin{minipage}[t]{0.55\textwidth}
        \vspace{0pt}
        \centering
        \begin{tabularx}{\linewidth}
            {@{}>{\raggedright\arraybackslash}Xcc@{}}
            \toprule
            \textbf{Method}
                & \textbf{SR (\%)}$\uparrow$
                & \textbf{MS}$\uparrow$ \\
            \midrule

            OpenVLA \citep{kim2024openvlaopensourcevisionlanguageactionmodel}
                & 1.5  & 6.1  \\
            RDT-1B \citep{liu2025rdt1bdiffusionfoundationmodel}
                & 5.3  & 17.7 \\
            $\pi_{0}$ \citep{black2026pi0visionlanguageactionflowmodel}
                & 8.2  & 24.0 \\
            $\pi_{0.5}$ \citep{intelligence2025pi05visionlanguageactionmodelopenworld}
                & 9.6  & 26.2 \\
            InternVLA-M1 \citep{chen2025internvlam1spatiallyguidedvisionlanguageaction}
                & 5.4  & 27.6 \\
            VLA-Adapter \citep{wang2025vlaadaptereffectiveparadigmtinyscale}
                & 4.4  & 24.3 \\
            $\pi_{0}$-FAST \citep{pertsch2025fastefficientactiontokenization}
                & 3.5  & 20.9 \\
            OpenVLA-OFT \citep{kim2025finetuningvisionlanguageactionmodelsoptimizing}
                & 9.1  & 24.1 \\
            StarVLA-OFT \citep{community2026starvlalegolikecodebasevisionlanguageaction}
                & 10.9 & 30.5 \\
            PUMA \citep{fang2026generalizableroboticmanipulationdynamic}
                & 17.2 & 35.0 \\

            \midrule
            \textbf{Ours}
                & \textbf{29.3} & \textbf{40.6} \\
            \bottomrule
        \end{tabularx}

        \par\vspace{4pt}
        {\footnotesize
        (a) DOMINO dynamic manipulation benchmark.\par}
    \end{minipage}%
    \hfill
    % ==================== Right table ====================
    \begin{minipage}[t]{0.43\textwidth}
        \vspace{0pt}
        \centering
        \begin{tabular*}{\linewidth}
            {@{\extracolsep{\fill}}lccc@{}}
            \toprule
            \textbf{Subtask}
                & \textbf{PUMA}
                & $\boldsymbol{\pi_{0.5}}$
                & \textbf{Ours} \\
            \midrule

            Bottle Return
                & 22\% & 20\% & \textbf{32\%} \\
            Can Grasp
                & 36\% & 92\% & \textbf{100\%} \\
            Stack Two
                & 24\% & 32\% & \textbf{38\%} \\
            Two Color Cue
                & 26\% & 22\% & \textbf{62\%} \\
            Can Placement
                & 42\% & 34\% & \textbf{52\%} \\
            Tray Swap
                & 14\% & 16\% & \textbf{38\%} \\
            Screen Cue
                & 20\% & 36\% & \textbf{90\%} \\
            Stack Three
                & 0\% & 6\% & \textbf{22\%} \\
            Motion Memory
                & 10\% & 52\% & \textbf{80\%} \\
            Color Memory
                & 12\% & 44\% & \textbf{86\%} \\

            \midrule
            \textbf{Average}
                & 20.6\% & 35.4\% & \textbf{60\%} \\
            \bottomrule
        \end{tabular*}

        \par\vspace{4pt}
        {\footnotesize
        (b) DOMINO-Long benchmark.\par}
    \end{minipage}

    \caption{
        \textbf{Evaluation on dynamic and long-horizon manipulation.}
        (a) Comparison on DOMINO,
        measured by SR and MS.
        (b) Per-subtask results on the 10 subtasks of DOMINO-Long.
    }
    \label{tab:dynamic_domino_results}
\end{table*}

\subsection{Evaluation On Dynamic Manipulation Benchmarks}
\label{sec:exp-domino-extension}

\paragraph{Dynamic manipulation on DOMINO.}
% BEGIN archived original: DOMINO
% {DOMINO evaluates interception and tracking of independently moving objects, testing whether the policy can use fresh visual feedback to act on changing target positions. We fine-tune on clean demonstrations from dynamic levels L1, L2, and L3 and evaluate in the clean L1 setting with predictable low-order motion. Table~\ref{tab:dynamic_domino_results} (a) reports complete-task success rate (SR) and Manipulation Score (MS), which measures execution quality beyond binary success. D$^2$-VLA achieves 28.5\% SR and 40.3 MS, the highest scores among the compared methods. It exceeds $\pi_{0.5}$ by 18.9 percentage points in SR and 14.1 points in MS, and PUMA, the strongest listed baseline, by 11.3 percentage points and 5.3 points, respectively. These gains support our joint dual-memory, dual-frequency design for dynamic manipulation. Training and inference settings are provided in Appendix~\ref{app:implementation}.}
% END archived original: DOMINO

\anlin{DOMINO tests D$^2$-VLA's responsiveness to changing observations through interception and tracking of independently moving objects, requiring correct target identification and continuous action adjustment. We fine-tune on clean demonstrations and evaluate on clean L1 with predictable low-order motion. Tab.~\ref{tab:dynamic_domino_results}(a) reports 29.3\% SR and 40.6 MS, which measures execution quality beyond binary success. D$^2$-VLA exceeds $\pi_{0.5}$ by 19.7 percentage points in SR and 14.4 points in MS, and PUMA by 12.1 percentage points and 5.6 points, respectively. These gains support the effectiveness of combining historical context with frequent visual feedback for dynamic manipulation. Training and inference details are in Appendix~\ref{app:implementation}.}

\anlin{\paragraph{Long-Horizon Dynamic Manipulation on DOMINO-Long.}
% We introduce DOMINO-Long to evaluate long-horizon decision making under dynamic interaction, where successful execution requires both retaining task-relevant information over extended temporal intervals and responding to continuously changing observations. DOMINO-Long comprises ten multi-stage manipulation tasks, each with 50 demonstrations, resulting in 500 demonstrations and 227,208 observation--action records in total, all used for training. Individual demonstrations contain 201--911 records, with an average length of 454.4. The demonstrations are generated using task-specific scripts and planning procedures and preserve observations from cue presentation through subsequent intermediate stages.

% The tasks are designed such that critical information, including previous object locations, presentation order, color cues, or motion direction, is observed earlier in the episode but must be recalled at a later stage to complete the manipulation. Representative examples include restoring a previous object arrangement, executing actions according to an earlier observed sequence, and selecting a moving object based on a cue that is no longer visible. By temporally separating the acquisition of task-relevant cues from the actions that depend on them, while simultaneously requiring interaction with moving objects, DOMINO-Long jointly evaluates long-horizon context retention and dynamic responsiveness. Complete-task success requires selecting the correct target or sequence and completing the manipulation, not cue recognition alone.

Tab.~\ref{tab:dynamic_domino_results}(b) reports 60\% average complete-task SR, exceeding $\pi_{0.5}$ (35.4\%) and PUMA (20.6\%) by 24.6 and 39.4 percentage points, respectively. These gains support the effectiveness of combining long-term memory with frequent visual updates for long-horizon dynamic manipulation. Observation/action interfaces are in Appendix~\ref{app:model_interfaces}; task definitions and evaluation settings are in Appendix~\ref{appendix:domino_long_dataset}.}

\subsection{Real-World Evaluation}
\label{sec:exp-real}

\begin{figure}[htbp]
    \centering
    \includegraphics[width=0.99\linewidth]{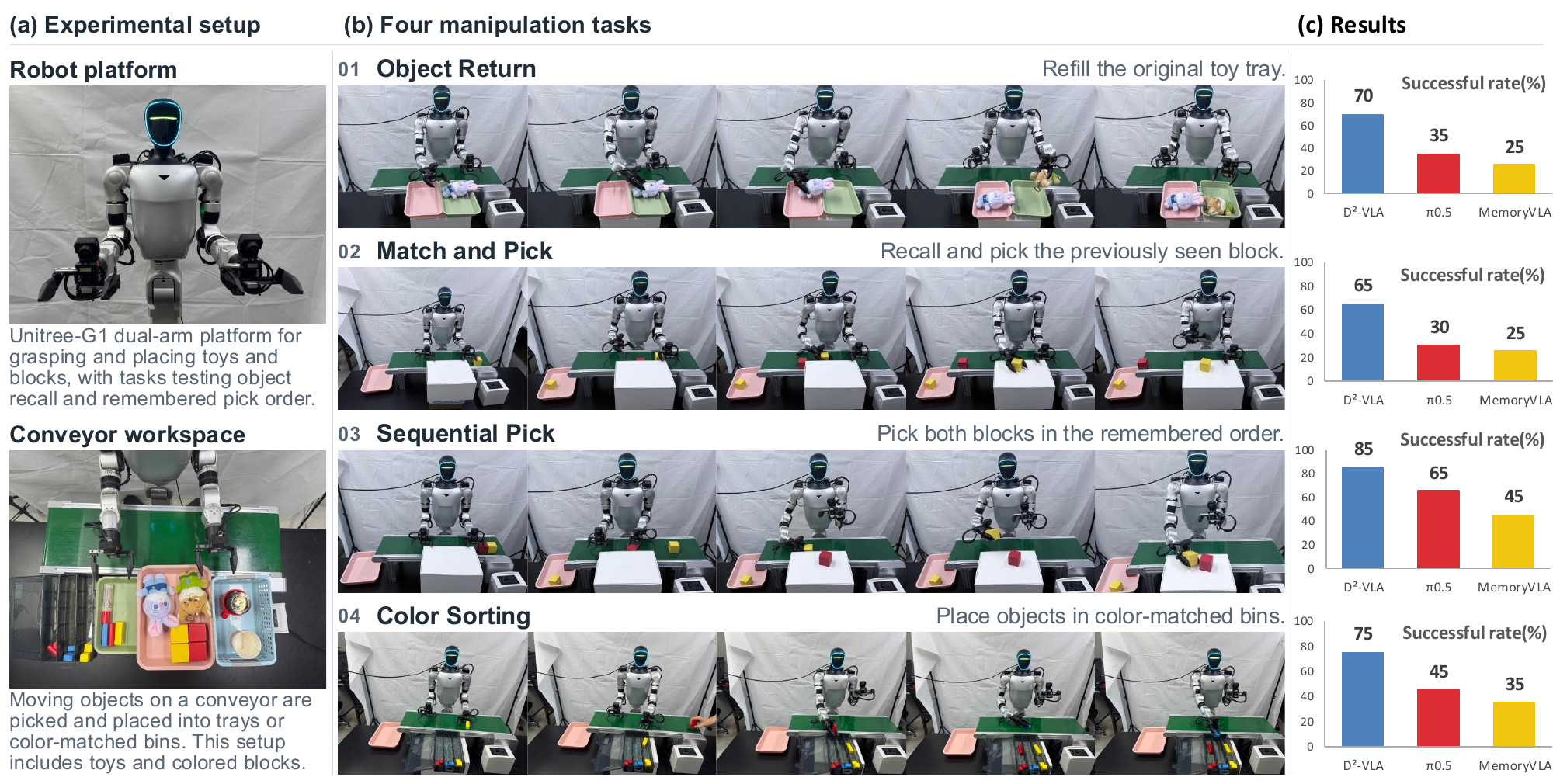}
    \caption{Long-horizon dynamic real-world evaluation on the Unitree G1D robot. Representative task executions and success-rate comparisons with $\pi_{0.5}$ and MemoryVLA.}
    \label{fig:Real_world_experiment}
\end{figure}

\paragraph{Task design.}

\anlin{We deploy D$^2$-VLA on a Unitree G1D for bimanual manipulation; the robot and observation/action interfaces are described in Appendix~\ref{app:real_platform}. Four static long-horizon tasks test ordered subgoals, object assignments, and memory of earlier spatial layouts. Four dynamic long-horizon tasks require decisions using current observations and task information retained from earlier in the episode. Together, they test whether $\mathrm{D}^{2}\text{-VLA}$ integrates long-term context with fresh visual feedback for multi-stage physical manipulation.
}
% Previous ten-task collection: 392 demonstrations and 460,190 observation--action records, with 20--54 demonstrations per task. Recompute for the selected eight tasks before reporting.

\paragraph{Evaluation and results.}
{We report complete-task SR, requiring the full instructed sequence. Fig.~\ref{fig:Real_world_experiment} presents dynamic-task examples and results. D$^2$-VLA improves over $\pi_{0.5}$ and MemoryVLA across tasks by varying margins, supporting retained context and frequent visual updates for real-world dynamic manipulation. Static-task results and additional examples are in Appendix~\ref{sec:real_world_tasks}.\par}
% TODO: Finalize per-task trial counts, matched initial conditions, motion settings,
% task-specific success criteria, episode limits, intervention rules, and deployment
% reset and feedback-rate settings in the appendix. Do not infer these from demonstration counts.

\subsection{Evaluation on Standard Static Benchmarks}
\label{sec:exp-simulation}
We evaluate D$^2$-VLA on LIBERO-Long~\cite{liu2023liberobenchmarkingknowledgetransfer} and RoboTwin~2.0~\cite{robotwin2} with the high-rate pathway disabled, retaining full-KV training and online dual-memory selection at inference. Tab.~\ref{tab:robotwin_libero_long} reports 97.5\% SR on LIBERO-Long, exceeding $\pi_{0.5}$ (92.4\%) and MemoryVLA (93.4\%) by 5.1 and 4.1 percentage points. On RoboTwin~2.0, D$^2$-VLA reaches 74.3\% SR, improving over $\pi_{0.5}$ (57.0\%) by 17.3 percentage points. These results support the usefulness of history-aware conditioning beyond dynamic scenes, without high-rate visual updates.

% Preamble

\begin{table*}[t]
\centering
% Shared typography and row spacing keep both panels the same height.
\footnotesize
\setlength{\tabcolsep}{3pt}
\renewcommand{\arraystretch}{1.15}

% ==================== Left: RoboTwin ====================
\begin{minipage}[t]{0.55\textwidth}
\vspace{0pt}
\centering

\begin{tabular*}{\linewidth}{@{\extracolsep{\fill}}lcc@{}}
\toprule
\textbf{Method}
& \textbf{Params (B)} $\downarrow$
& \textbf{SR (\%)} $\uparrow$ \\
\midrule

Diffusion Policy~\citep{chi2024diffusionpolicyvisuomotorpolicy}
    & 0.1 & 28.0 \\

ACT~\citep{zhao2023learningfinegrainedbimanualmanipulation}
    & 0.1 & 29.7 \\

DP3~\citep{ze20243ddiffusionpolicygeneralizable}
    & 0.3 & 55.2 \\

$\pi_0$~\citep{black2026pi0visionlanguageactionflowmodel}
    & 3.2 & 46.4 \\

FlowPolicy~\citep{zhang2024flowpolicyenablingfastrobust}
    & 0.3 & 41.0 \\

RDT-1B~\citep{liu2025rdt1bdiffusionfoundationmodel}
    & 1.7 & 34.5 \\

SeedPolicy~\citep{gui2026seedpolicyhorizonscalingselfevolving}
    & 0.2 & 42.8 \\

Turbo-VLA~\citep{xie2026turbovla}
    & 0.4 & 60.2 \\

$\pi_{0.5}$~\citep{intelligence2025pi05visionlanguageactionmodelopenworld}
    & 3.4 & 57.0 \\

StarVLA-$\alpha$~\citep{ye2026starvlaalphareducingcomplexityvisionlanguageaction}
    & 3.8 & 50.3 \\

\midrule

\textbf{D$^{2}$-VLA (Ours)}
    & \textbf{3.5} & \textbf{74.3} \\

\bottomrule
\end{tabular*}

\par\vspace{4pt}
{\footnotesize (a) RoboTwin 2.0.\par}
\end{minipage}%
\hfill
% ==================== Right: LIBERO-LONG ====================
\begin{minipage}[t]{0.43\textwidth}
\vspace{0pt}
\centering

\begin{tabular*}{\linewidth}{@{\extracolsep{\fill}}lc@{}}
\toprule
\textbf{Method}
& \textbf{SR (\%)} $\uparrow$ \\
\midrule

CogACT~\citep{li2024cogactfoundationalvisionlanguageactionmodel}
    & 53.2 \\

OpenVLA~\citep{kim2024openvlaopensourcevisionlanguageactionmodel}
    & 53.7 \\

$\pi_0$-FAST~\citep{pertsch2025fastefficientactiontokenization}
    & 60.2 \\

CronusVLA~\citep{li2025cronusvlaefficientrobustmanipulation}
    & 68.7 \\

CoT-VLA~\citep{zhao2025cotvlavisualchainofthoughtreasoning}
    & 69.0 \\

SmolVLA~\citep{shukor2025smolvlavisionlanguageactionmodelaffordable}
    & 77.0 \\

$\pi_0$~\citep{black2026pi0visionlanguageactionflowmodel}
    & 85.2 \\

GR00T-N1~\citep{nvidia2025gr00tn1openfoundation}
    & 90.6 \\

$\pi_{0.5}$~\citep{intelligence2025pi05visionlanguageactionmodelopenworld}
    & 92.4 \\

MemoryVLA~\citep{shi2026memoryvlaperceptualcognitivememoryvisionlanguageaction}
    & 93.4 \\
    
\midrule

\textbf{D$^{2}$-VLA (Ours)}
    & \textbf{97.5} \\

\bottomrule
\end{tabular*}

\par\vspace{4pt}
{\footnotesize (b) LIBERO-Long.\par}
\end{minipage}

\caption{
\textbf{Performance comparison on static manipulation benchmarks.}
(a) Model size and average success rate on RoboTwin 2.0,
with all methods trained and evaluated exclusively on the clean setting.
(b) Average success rate on LIBERO-Long.
}
\label{tab:robotwin_libero_long}
\end{table*}

% \paragraph{Long-horizon manipulation on LIBERO-Long.}
% {The ten-task LIBERO-Long suite provides a complementary evaluation of multi-stage, language-conditioned manipulation. We use it to assess whether the proposed temporal conditioning also supports extended task execution outside the dynamic-interaction settings of DOMINO and DOMINO-Pro. Complete-task SR, averaged across tasks, is reserved in Table~\ref{tab:exp-simulation}. Training data, observation inputs, episode limits, and trial counts are [TBD].\par}
% \emph{Results: [TBD].}

% \paragraph{Supplementary evaluation on RoboTwin~2.0.}
% {RoboTwin~2.0 supplies a supplementary evaluation of bimanual manipulation across varied task configurations. Its role is to assess the broader applicability of the policy beyond the memory-focused and dynamic tasks emphasized above. We reserve aggregate complete-task SR in Table~\ref{tab:exp-simulation}; the evaluated task set, training data, clean/randomized conditions, trial counts, and results remain [TBD].\par}

\subsection{Ablation Studies}
\label{sec:exp-analysis}

% Start the first paragraph after the heading before installing wraptable's paragraph hook.
\edef\TableThreeIntextsep{\the\intextsep}
\setlength{\intextsep}{2pt}
\leavevmode
\begin{wraptable}{r}{0.46\textwidth}
\centering
\footnotesize
\setlength{\tabcolsep}{3pt}
\renewcommand{\arraystretch}{1.15}
\begin{tabular*}{\linewidth}{@{\extracolsep{\fill}}lccc@{}}
\toprule
\textbf{Model} & \textbf{DM} & \textbf{DF} & \textbf{SR (\%)}\\
\midrule
$\pi_{0.5}$ & -- & -- & 9.6\\
$+\,$ Causal Training & \checkmark & -- & 16.1\\
\textbf{D$^{2}$-VLA} & \checkmark & \checkmark & \textbf{29.3}\\
\bottomrule
\end{tabular*}
\caption{Component ablation on DOMINO. DM: dual-memory; DF: dual-frequency.}
\label{tab:main_ablation}
\label{tab:ablation_components}
\end{wraptable}
\color{black}
\noindent\textbf{Component ablation.}
We progressively add causal training with dual-memory selection and then dual-frequency control, including the adapter, to $\pi_{0.5}$ on DOMINO. The first stage uses separate VLM and expert history reads without high-rate refinement; the second adds fast memory between VLM refreshes. Tab.~\ref{tab:ablation_components} shows that the first stage raises SR from $9.6\%$ to $16.1\%$ (+6.5 percentage points), and dual-frequency control further increases it to $29.3\%$ (+13.2 points). These gains support complementary benefits from historical context and intermediate visual updates. Additional ablations on historical KV retention and update periods are reported in Appendices~\ref{app:selection_analysis} and~\ref{app:memory_sensitivity}, respectively. Memory and latency analysis can be seen in Appendix~\ref{app:kv_cache_tradeoff}.
\par
\color{black}
\setlength{\intextsep}{\TableThreeIntextsep}

\section{Conclusion}

We presented D$^2$-VLA, combining causal KV reuse, consumer-specific history selection, and adapter-based dual-frequency control to retain context and use fresh observations without a VLM pass at every replanning step. We introduced DOMINO-Long to test long-horizon memory and dynamic responsiveness. D$^2$-VLA achieves 29.3\% SR on DOMINO and 60.0\% on DOMINO-Long, improves dynamic real-world manipulation, and extends memory-based design to static benchmarks, demonstrating the substantial benefits of combining historical context with timely visual feedback.

\subsection*{AI use statement}

In this work, we used generative AI tools solely for language editing and proofreading, including improving grammar, readability, and overall clarity of the manuscript. We did not use generative AI tools for generating research ideas, designing methods, conducting experiments, analyzing results, producing scientific claims, or creating research artifacts. All AI-assisted edits were carefully reviewed and revised by the authors to ensure accuracy and consistency with the intended content. The authors take full responsibility for the final content of this work, including all text, claims, and artifacts.

\subsection*{Reproducibility statement}

To facilitate reproducibility, we are willing to provide anonymous access to the source code, model weights, and the training and inference data used in our experiments for reproducibility. Detailed descriptions of the proposed method, experimental settings, and implementation details are provided in the main paper and appendix. These resources enable independent verification of our results and support further research on long-horizon vision-language-action models.

\bibliography{iclr2027_conference}
\bibliographystyle{iclr2027_conference}

\newpage

\appendix

\section{Implementation and Training Details}
\label{app:implementation}

% This appendix specifies the Full-Block implementation of D$^2$-VLA used for DOMINO, DOMINO-Long, and real-world evaluation. All three settings use full fine-tuning from $\pi_{0.5}$ Base, four slow blocks, and two fast blocks. These capacities include the current block, leaving up to three historical slow observations and one historical fast observation. We denote the predicted action horizon by $H$ and the executed chunk size by $E$; the reported configuration uses $E=H$. The online history selector is applied only at inference; training uses uncompressed conditioning memories. Simulation evaluation protocols are given in Appendix~\ref{app:simulation_protocols}, and physical setup and evaluation are given in Appendix~\ref{sec:real_world_tasks}.

\anlin{
This appendix provides the implementation and training details of D$^2$-VLA. We first describe the model architecture, observation and action interfaces. We then detail the implementation of the key VLA components, including the causal modeling and cache conventions, dynamic feature adaptation, model optimization and inference. Finally, we summarize the experiment-specific configurations for DOMINO, DOMINO-Long, and the real-world tasks, together with the corresponding simulation protocols, real-world evaluation settings, memory analysis, and ablation studies. 
}

\subsection{Model Architecture and Observation--Action Interfaces}
\label{app:model_interfaces}

\paragraph{Backbone and token layout.}
% The VLM is initialized from the PaliGemma-3B backbone~\cite{paligemma} of $\pi_{0.5}$, comprising a SigLIP~\cite{siglip} vision encoder and a Gemma language backbone. The language backbone uses the Gemma-2b~\cite{gemma2b} configuration; the shared action expert uses Gemma-300m~\cite{gemma300m}. These configuration names refer to the language backbone and action expert, not the parameter count of the complete VLM. Both Transformer stacks have 18 layers, 8 query heads, one KV head, and head dimension of 256; their hidden widths are 2048 and 1024, and their feed-forward widths are 16,384 and 4096, respectively. The shared SigLIP encoder uses $14\times14$ patches, mapping each $224\times224$ image to 256 visual tokens projected to width of 2048. With three camera views and a maximum instruction length of 200 tokens, one observation block allocates 768 visual slots and 200 instruction slots, giving 968 physical positions; the number of valid tokens can be smaller because unused instruction slots and unavailable views are masked. The instruction is repeated in each observation block. In our model, task-only tokenization excludes state from the VLM prompt. The normalized state is padded to 32 dimensions and independently projected to one 1024-dimensional action-expert token, which is not stored in the observation cache.

The VLM is initialized from the PaliGemma-3B backbone~\cite{paligemma} used in $\pi_{0.5}$. It consists of a SigLIP vision encoder~\cite{siglip} and a Gemma language backbone. The language backbone follows the Gemma-2B configuration, while the action expert is based on Gemma-300M. Both Transformer stacks contain 18 layers, with 8 query heads, 1 KV head, and a head dimension of 256. Their hidden dimensions are 2048 and 1024, respectively.

The SigLIP~\cite{siglip} encoder processes each $224\times224$ image using $14\times14$ patches and produces 256 visual tokens. With three camera views, each observation contains 768 visual tokens. We allow up to 200 instruction tokens, resulting in a maximum of 968 token positions per observation block. Unused instruction positions and unavailable camera views are masked. The same instruction is repeated for each observation block.

The robot state is handled separately from the VLM input. The normalized state is padded to 32 dimensions and projected into a single 1024-dimensional token for the action expert. This state token is not stored in the visual-language cache.

\paragraph{Embodiment-specific inputs and actions.}
% The simulation interface uses a head view and two wrist views. DOMINO-Long recordings have original image resolution $640\times480$ and 14-dimensional ALOHA state/action vectors: 12 arm-joint dimensions and two gripper dimensions. The Unitree interface instead uses 16 dimensions: 14 arm-joint dimensions and two gripper dimensions. Both are padded to 32 dimensions inside the model. For the real robot, the left and right head images are combined into one model view, while the two wrist images remain separate; the exact construction is described in Appendix~\ref{app:real_platform}.

% For each supervised branch, joint targets are expressed relative to that branch's current state,
% \begin{equation}
%  \widetilde A_{t_j,h,d}=A^{\mathrm{abs}}_{t_j+h,d}-s_{t_j,d},
%  \label{eq:app_relative_action}
% \end{equation}
% where $t_j$ is the current time of branch $j$, $h=0,\ldots,H-1$ is the action offset, and $d$ indexes an arm joint. Here $s_{t_j,d}$ is the physical joint state before normalization. Gripper targets remain absolute. This conversion precedes quantile normalization. At inference, the branch state is added back after denormalization, and padded dimensions are discarded. Normalization statistics are computed separately for each training dataset. Invalid action steps are masked in the training loss as described in Appendix~\ref{app:training_details}.

The simulation setup uses three camera views: one head camera and two wrist cameras. For DOMINO-Long, the original images are recorded at $640\times480$ resolution. The ALOHA state and action vectors contain 14 dimensions, including 12 arm-joint dimensions and two gripper dimensions. The Unitree robot uses 16-dimensional state and action vectors, consisting of 14 arm-joint dimensions and two gripper dimensions. For a unified model interface, both representations are padded to 32 dimensions. In the real-world setup, the left and right head-camera images are combined into a single model view, while the two wrist-camera views are kept separate. Additional details are provided in Appendix~\ref{app:real_platform}.

For training, arm-joint actions are represented relative to the current robot state:
\begin{equation}
\widetilde A_{t_j,h,d}
=
A^{\mathrm{abs}}_{t_j+h,d}
-
s_{t_j,d},
\label{eq:app_relative_action}
\end{equation}
where $t_j$ denotes the current step of branch $j$, $h=0,\ldots,H-1$ is the action offset, and $d$ indexes an arm joint. The gripper actions remain in absolute coordinates. Relative joint actions are computed before quantile normalization. During inference, the current joint state is added back after denormalization to recover absolute joint commands, and padded dimensions are removed before execution. Normalization statistics are computed separately for each training dataset, and invalid action steps are excluded from the training loss as described in Appendix~\ref{app:training_details}.

\subsection{Causal Modeling and Cache Conventions}
\label{app:layouts}
\label{app:causal_details}

\paragraph{Observation-level causality.}
% Visual and instruction tokens within an observation block attend bidirectionally, while attention across observation blocks is causal. Training windows contain four observation slots: the current observation and up to three historical observations. Missing early observations are left-padded and masked, and histories never cross episode boundaries. The action expert reads valid observation tokens together with the current state and noisy action tokens. For four observation blocks, the block-level visibility is
% \begin{equation}
% \begin{array}{c|cccccc}
%  &B_1&B_2&B_3&B_4&z_t^s&A_t^\tau\\ \hline
%  B_1&1&0&0&0&0&0\\
%  B_2&1&1&0&0&0&0\\
%  B_3&1&1&1&0&0&0\\
%  B_4&1&1&1&1&0&0\\
%  z_t^s&1&1&1&1&1&0\\
%  A_t^\tau&1&1&1&1&1&1
% \end{array}.
% \label{eq:app_causal_visibility}
% \end{equation}
% Here $B_1,\ldots,B_4$ index the observation slots chronologically within the window, rather than absolute environment times as in $B_j$ in Section~\ref{sec:causal_modeling}.
% The token $z_t^s$ is the projected current-state token, and $A_t^\tau$
% denotes the noisy action tokens at flow time $\tau$, defined in
% Appendix~\ref{app:training_details}. Each entry denotes a token
% submatrix before applying validity masks. This matrix specifies information visibility, not a single shared Transformer: the VLM encodes observation blocks, and the expert processes state/action tokens against their layerwise KV.

% \paragraph{Observation-level causal attention.}
We organize each training sample as a temporal window containing the current observation and up to three preceding observations. Tokens within the same observation block interact bidirectionally, whereas attention across observation blocks is causal: each block can attend to itself and all earlier blocks, but not to future ones. Missing history at the beginning of an episode is padded and masked, and temporal windows never cross episode boundaries. This follows the causal modeling scheme introduced in Section~\ref{sec:causal_modeling}.

The action expert receives the encoded observation history together with the current state token $z_t^s$ and the noisy action tokens $A_t^\tau$. The state token can attend to all valid observation blocks, while the action tokens can additionally attend to the current state and to one another. For a temporal window containing $N$ observation blocks, the visibility pattern can be written as
\begin{equation}
\mathcal{V}(B_i)
=
\{B_j \mid j \leq i\},
\qquad
\mathcal{V}(z_t^s)
=
\{B_1,\ldots,B_N,z_t^s\},
\qquad
\mathcal{V}(A_t^\tau)
=
\{B_1,\ldots,B_N,z_t^s,A_t^\tau\},
\label{eq:app_causal_visibility}
\end{equation}
where $\mathcal{V}(\cdot)$ denotes the set of tokens visible to a given block or token group. In our implementation, $N=4$, with $B_1,\ldots,B_4$ ordered from the oldest to the most recent observation, rather than indexing absolute environment times as in Section~\ref{sec:causal_modeling}. The noisy action tokens $A_t^\tau$ follow the flow-matching formulation defined in Appendix~\ref{app:training_details}. The VLM encodes the observation blocks, while the action expert operates on their layer-wise KV representations together with the state and action tokens.

\paragraph{Persistent slow-memory cache.}
The slow-memory queue $Q_{\mathrm{slow}}$ stores complete layer-wise KV blocks in temporal order. At each slow update, the retained blocks are concatenated along the token dimension to form the current slow-memory context $\mathcal{K}_t$. The concatenation is performed separately for keys and values at every Transformer layer while preserving the temporal order of the blocks:
\begin{equation}
\mathcal{K}_t
=
\operatorname{Concat}_{\mathrm{KV}}
\left(
Q_{\mathrm{slow}}^{t}
\right).
\label{eq:queue_cache_correspondence}
\end{equation}

Before encoding a new slow observation at step $t$, the queue retains at most the most recent $N_{\mathrm{slow}}-1$ historical blocks. Let
$Q_{\mathrm{slow}}^{t^-}$ denote the queue state immediately before the current update. The historical context, current KV block, and updated slow-memory queue are given by
\begin{equation}
\begin{aligned}
Q_{\mathrm{slow}}^{<t}
&=
\operatorname{P}_{N_{\mathrm{slow}}-1}
\left(
Q_{\mathrm{slow}}^{t^-}
\right),\\
\mathcal{K}_{<t}
&=
\operatorname{Concat}_{\mathrm{KV}}
\left(
Q_{\mathrm{slow}}^{<t}
\right),\\
\mathcal{K}_c(t)
&=
\Phi_{\mathrm{KV}}
\left(
I_t,\ell;
\operatorname{Read}_{\mathrm{VLM}}
\left(
\mathcal{K}_{<t}
\right)
\right),\\
Q_{\mathrm{slow}}^{t}
&=
Q_{\mathrm{slow}}^{<t}
\oplus
\mathcal{K}_c(t).
\end{aligned}
\label{eq:incremental_cache}
\end{equation}
Here, $\operatorname{P}_{k}(\cdot)$ preserves up to the most recent $k$ complete blocks, $\operatorname{Concat}_{\mathrm{KV}}(\cdot)$ concatenates their keys and values separately along the token dimension, and $\oplus$ denotes temporal concatenation. In our implementation, $N_{\mathrm{slow}}=4$, so the queue stores the current block together with up to three preceding slow observations. The history read follows the selection rule defined in Section~\ref{sec:dual_memory_selection}, while the temporal ordering follows the causal formulation in Section~\ref{sec:causal_modeling}.

Importantly, historical blocks are encoded only once and are not recomputed at every slow update. The history selector affects only the temporary read view used by $\Phi_{\mathrm{KV}}$; complete retained blocks remain stored in $Q_{\mathrm{slow}}$. Consequently, the slow-memory cache is not equivalent to repeatedly encoding a fresh four-frame window, since each retained block may already contain information propagated from earlier observations.

% \paragraph{Raw KV and positions.}
% Both slow and fast memories store keys before rotary position encoding, together with their values; rotary encoding is not applied to values. Each read view retains the original block/token order and assigns compact positions to valid tokens; rotary embeddings are applied when the keys are consumed. Invalid camera slots and instruction padding remain masked. Position reassignment does not recompute the stored K/V contents. The online selector uses this same repeated Image--Task layout and raw-KV convention, rather than a separate task-prefix layout.

\paragraph{KV storage and positional encoding.}
Both the slow and fast memories store raw keys and values before rotary positional encoding is applied. Rotary embeddings are applied to the keys only when cached features are read, while the values remain unchanged. During each read, valid tokens preserve their original temporal and within-block order but are reassigned compact position indices. Masked camera tokens and padded instruction tokens remain invalid throughout this process. Thus, position reassignment affects only the positional encoding used at read time and does not modify the stored KV features. The online history selector follows the same Image--Task block layout and raw-KV storage format as the memory queues, ensuring consistent representations across caching, selection, and retrieval.

\subsection{Online Consumer-Specific Historical KV Selection}
\label{app:selection_details}

\paragraph{Separate online statistics.}
Historical KV selection is applied only at inference and introduces no trainable parameters.
Let $g\in\{\mathrm{p},\mathrm{s},\mathrm{f}\}$ index VLM prefill, slow-branch action denoising, and fast-branch action denoising, respectively.
Each stage maintains its own score tensor independently for each sample.
Prefill and slow-denoising scores are associated with complete slow-memory blocks; fast-denoising scores are associated with the refined-block FIFO.
The two action branches share expert weights, but maintain separate history-importance statistics. Within each consumer, selected token indices are shared across layers.
In Eq.~\ref{eq:memory_views}, $\operatorname{Read}_{\mathrm{VLM}}$ denotes the prefill read ($g=\mathrm{p}$), while $\operatorname{Read}_{\mathrm{act}}$ denotes the slow-denoising read ($g=\mathrm{s}$).
At intermediate replanning steps in Eq.~\ref{eq:dual_rate_condition}, $\operatorname{Read}_{\mathrm{act}}$ instead uses fast-denoising scores ($g=\mathrm{f}$) to select from fast memory.
The two action paths share the same read operator notation, but use separate source memories and importance statistics.

For each sample, let $i$ index a token in the stage's full stored memory and $l$ index a Transformer layer.
We denote its accumulated importance by $h_{g,l,i}$.
The attention statistic $\bar a_{g,l,i}$ averages the existing attention probabilities over heads and valid queries in the current call.
For prefill, queries are the valid current observation tokens; for denoising, they include the current state token and action tokens. Denoising statistics are additionally averaged over flow-integration steps.
With score-retention coefficient $\rho\in[0,1)$, fixed at 0.2 in the reported experiments, the update and cross-layer selection score are
\begin{equation}
 \begin{aligned}
 h_{g,l,i}^{+}
 &=
 \begin{cases}
 \rho h_{g,l,i}+(1-\rho)\bar a_{g,l,i},
      & i\text{ is observed in the current read},\\
 h_{g,l,i},&\text{otherwise},
 \end{cases}\\
 u_{g,i}&=\max_l h_{g,l,i}.
 \end{aligned}
 \label{eq:app_online_importance}
\end{equation}
Here $+$ denotes the score state after the current call, and $u_{g,i}$ is computed from the score state available before each read.
The current forward pass therefore updates scores for subsequent reads, not its own selection.
Unobserved tokens are not decayed.
New score slots are zero-initialized, scores are removed with evicted blocks, and all streaming state is reset between episodes.
Statistics are obtained from the forward passes already used for encoding or denoising; selection does not require an additional full-memory scoring pass, an offline attention prior, or a K/V-magnitude term.

\paragraph{Per-block visual budgets and shared indices.}
Each stage supports a separately configurable visual-token budget for each historical block, specified either as a token count or as a retention ratio. In the reported experiments, all active consumers use the same retention ratio, while maintaining independent attention statistics. The main results use a historical visual-token retention ratio of 0.7.
An explicit token budget takes precedence over the ratio. Without an explicit budget, the per-block visual budget is the ceiling of the retention ratio times the number of allocated visual slots; masked camera slots remain invalid and their quota is not reassigned.
The budget is allocated across cameras: the implementation first reserves up to four tokens per camera when the budget permits, then distributes the remaining capacity proportionally.
If the total budget is smaller, the initial allocation proceeds round-robin.
Let $\kappa_{g,c}$ be the resulting quota for camera $c$ in each historical block at stage $g$.
At time $t$, let $\mathcal{V}_{t,j,c}^{g}$ denote the valid visual-token indices from camera $c$ in a retained historical block with observation time $j<t$, within that stage's source memory.
We define $\mathcal{J}_{t,j,c}^{g}$ as the set of visual-token indices selected from this group:
\begin{equation}
 \mathcal{J}_{t,j,c}^{g}
 =\operatorname{TopK}_{i\in\mathcal{V}_{t,j,c}^{g}}
 \!\left(u_{g,i},\min\!\left(\kappa_{g,c},|\mathcal{V}_{t,j,c}^{g}|\right)\right).
 \label{eq:selection_rule}
\end{equation}
The operator $\operatorname{TopK}$ returns the highest-scoring indices.
Thus, selection is performed separately within each historical block and camera, rather than globally across the entire history.
Selected positions are sorted back into their original order; equal-score ties favor earlier source positions.
The layerwise maximum in Eq.~\ref{eq:app_online_importance} yields one index set shared by keys and values at every layer, with separate selection for each batch element.
Padding slots retained for static tensor shapes keep their validity masks and are not treated as valid selected tokens.

\paragraph{Protected tokens and read lifetime.}
\begin{wraptable}{r}{0.56\linewidth}
\vspace{-1pc}
\centering
% \vspace{-.3pc}
% \caption{\small Online KV read-view configuration.}
\caption{Online KV selection settings for VLM prefill, slow-branch denoising, and fast-branch denoising. Source, Current, and Ret. denote historical source, current-block protection, and visual-token retention ratio, respectively.}
\label{tab:selection_settings}
\vspace{-.7pc}
\small
\setlength{\tabcolsep}{3pt}
\renewcommand{\arraystretch}{1.08}

\begin{tabular*}{\linewidth}
{@{\extracolsep{\fill}}lccc@{}}
\toprule
\textbf{Consumer}
& \textbf{Source}
& \textbf{Current}
& \textbf{Ret.} \\
\midrule

VLM prefill
& Slow
& -- 
& 0.7 \\

Slow denoising
& Slow
& Yes
& 0.7 \\

Fast denoising
& Fast FIFO
& Yes
& 0.7 \\

\midrule
$\rho$
& \multicolumn{3}{c}{0.2 (fixed)} \\

\bottomrule
\end{tabular*}

\vspace{-1.4pc}
\end{wraptable}

The read operator combines the selected visual KV entries with all
instruction tokens from each historical block while preserving their
validity masks and temporal order. Instruction tokens are always retained
and do not count toward the visual-token budget. During VLM prefill, all
cached blocks are treated as history and are eligible for visual-token
selection, while the current observation is encoded in full.

For action denoising, the newest block is always kept complete, and only
earlier blocks are compressed. Thus, in the fast branch, compression is
applied only to preceding refined blocks when available, while the current
refined block remains intact. Selection affects only the temporary read
view; the complete KV blocks remain stored in memory as described in
Appendix~\ref{app:causal_details}.

Each denoising read view is constructed once per expert call and remains
fixed throughout the corresponding flow solve. Attention statistics
collected during the current call are used only to update importance scores
for subsequent reads. The selection mechanism operates within retained
historical blocks and does not restrict the action expert to a fixed number
of recent observations.
\subsection{Full-Block Adapter and Dual-Rate Inference}
\label{app:dual_details}

\paragraph{Update schedule.}
Using the notation of Section~\ref{sec:dual_frequency}, the slow-update
period $V$ is an integer multiple of the executed chunk size $E$.
The history-sampling interval matches this period, as specified in
Eq.~\ref{eq:elapsed_phase}.
With environment steps counted from episode start, the latest
slow-update time and normalized elapsed time are
\begin{equation}
 t_s=V\left\lfloor t/V\right\rfloor,\qquad
 \delta_t=\frac{t-t_s}{V}.
 \label{eq:dual_rate_schedule}
\end{equation}
A slow refresh occurs at $t=t_s$; other replanning calls use the
fast pathway. This elapsed-time input is distinct from flow time.

\paragraph{Current-observation encoding.}
The high-rate pathway applies the shared image encoder to the current
three-view observation, producing $E_{\mathrm{img}}(I_t)$ with 768 visual tokens of
width 2048. The observation encoder in Eq.~\ref{eq:fast_feature_update}
expands as
\begin{equation}
 \operatorname{Enc}_{\psi}(E_{\mathrm{img}}(I_t),\delta_t)
 =\operatorname{Tr}_{\psi}\!\left([
 \operatorname{Proj}(E_{\mathrm{img}}(I_t))+E_{\mathrm{cam}};
 \operatorname{MLP}_{\delta}(\delta_t)+e_{\delta}]
 \right).
 \label{eq:app_observation_encoder}
\end{equation}
Here $\operatorname{Proj}$ maps visual features to width 1024,
$E_{\mathrm{cam}}$ supplies learned camera embeddings, and
$\operatorname{MLP}_{\delta}$ with the learned embedding $e_{\delta}$
produces one temporal-offset token; the semicolon denotes token
concatenation. The context Transformer $\operatorname{Tr}_{\psi}$
has two layers, eight heads, and MLP width 4096, and processes the
resulting 769 positions with validity masks.
The pooler uses 32 learned base queries to attend to the encoded
context, followed by normalization and an MLP, producing the compact queries $\mathcal{Z}_t$ in Eq.~\ref{eq:fast_feature_update}.
The base queries are learned parameters, whereas $\mathcal{Z}_t$ depends on
the current observation. The encoder and pooler are shared across
VLM layers.

\paragraph{Layerwise full-block refinement.}
For layer $l$, separate normalization and projection operations map
$\mathcal{Z}_t$ and the raw keys and values of $\mathcal{K}_c(t_s)$ to editor width
$d_e=1024$, yielding $\widetilde Q_{t,l}$,
$\widetilde K_l$, and $\widetilde V_l$.
The two-stage editor first reads the slow block and then decodes
the routed features back to every block position:
\begin{equation}
 \begin{aligned}
 R_{t,l}
 &=\operatorname{softmax}\!\left(
 \frac{\widetilde Q_{t,l}\widetilde K_l^\top}{\sqrt{d_e}}
 +\Lambda_{t_s}\right)\widetilde V_l,\\
 D_{t,l}
 &=\operatorname{softmax}\!\left(
 \frac{\widetilde K_lR_{t,l}^{\top}}{\sqrt{d_e}}
 \right)R_{t,l},\\
 [\Delta K_t^{(l)},\Delta V_t^{(l)}]
 &=\operatorname{Head}_{l}(D_{t,l}).
 \end{aligned}
 \label{eq:app_editor_refinement}
\end{equation}
Here $\Lambda_{t_s}$ is an additive validity mask, equal to zero
at valid slow-block positions and negative infinity at invalid
positions. The softmax operations normalize over slow-block
positions and routing queries, respectively.
The layer-specific head applies layer normalization and a
$1024\rightarrow4096\rightarrow512$ MLP with GELU, producing separate
256-dimensional key and value residuals. Residuals at invalid
positions are set to zero.
The residuals are added to the original, unprojected KV through
layerwise gates:
\begin{equation}
 \begin{aligned}
 K_{t,\mathrm{ref}}^{(l)}
 &=K_c^{(l)}(t_s)+\tanh(g_K^{(l)})\Delta K_t^{(l)},\\
 V_{t,\mathrm{ref}}^{(l)}
 &=V_c^{(l)}(t_s)+\tanh(g_V^{(l)})\Delta V_t^{(l)}.
 \end{aligned}
 \label{eq:app_gated_kv_update}
\end{equation}
Here $K_c^{(l)}(t_s)$ and $V_c^{(l)}(t_s)$ are the layer-$l$ keys and
values of the complete slow anchor $\mathcal{K}_c(t_s)$.
The refined pairs across all layers form $\widetilde{\mathcal{Z}}_t$
in Eq.~\ref{eq:fast_feature_update}; $\operatorname{F}_{\psi}$ collects
their predicted residuals, while $G_{\psi}$ collects the effective
gates $\tanh(g_K^{(l)})$ and $\tanh(g_V^{(l)})$, broadcast over token
and feature dimensions. The final residual projections are
zero-initialized, and the raw scalar gates $g_K^{(l)}$ and $g_V^{(l)}$
are initialized to $\operatorname{arctanh}(0.1)$.
Refinement therefore starts as an identity update.
The 32 routing queries are an internal bottleneck, not the number of
output KV tokens: the output preserves all 968 physical positions.
The adapter always reads the full canonical slow block, even when
the VLM or expert uses compressed read views. It has no separate
task-text input; task information is carried by the slow Image--Task
KV. Current state enters the expert independently.

\paragraph{Slow and fast conditions.}
The slow cache $\mathcal{K}_{t_s}$ retains at most $N_{\mathrm{slow}}=N=4$ complete
blocks, including the latest block. At a slow refresh, the expert
condition combines the complete $\mathcal{K}_c$ with selected history as in
Eq.~\ref{eq:memory_views}. At an intermediate update, each $\widetilde{\mathcal{Z}}_t$ is
generated independently from $\mathcal{K}_c(t_s)$. The ordered queue $Q_{\mathrm{fast}}$
stores these complete refined blocks. Let $\mathcal{M}_t^f$ denote
its layerwise temporal concatenation at time $t$, and
$\mathcal{M}_{<t}^f$ the concatenation of its retained historical
part $Q_{\mathrm{fast}}^{<t}$. Their updates are
\begin{equation}
 \begin{aligned}
 \mathcal{M}_{t_s}^{f}&=\varnothing,\\
 \mathcal{M}_{<t}^{f}
 &=\operatorname{P}_{N_{\mathrm{fast}}-1}(\mathcal{M}_{t-E}^{f}),\\
 \mathcal{M}_t^{f}
 &=\mathcal{M}_{<t}^{f}\oplus \widetilde{\mathcal{Z}}_t,\qquad t>t_s,
 \end{aligned}
 \label{eq:fast_fifo}
\end{equation}
where $\operatorname{P}_k$ preserves up to the latest $k$ complete
blocks and $N_{\mathrm{fast}}=2$. The recursion applies only at intermediate replanning calls $t=t_s+kE<V+t_s$, with integer $k\geq1$. The FIFO is reset to $\mathcal{M}_{t_s}^{f}=\varnothing$ at each slow refresh, so the
first fast update has no historical fast block.
At intermediate replanning calls, the expert uses the fast condition
in Eq.~\ref{eq:dual_rate_condition}, protecting $\widetilde{\mathcal{Z}}_t$
while selecting historical visual tokens with independent
fast-denoising statistics. Slow and fast caches are not concatenated
into a single expert condition. The next slow refresh clears the
fast FIFO and its scores.

\begin{algorithm}[t]
\caption{Dual-rate inference with online historical KV selection}
\label{alg:dual_frequency}
\begin{algorithmic}[1]
\Require Action horizon $H$, executed chunk size $E=H$, slow period $V$
\Require Slow capacity $N_{\mathrm{slow}}=N=4$, fast capacity $N_{\mathrm{fast}}=2$
\State Initialize empty slow/fast memories and consumer-specific scores
\For{each replanning call at $t$ within an episode}
 \State Observe current images $I_t$, state $s_t$, and instruction $\ell$
 \State Set $t_s=V\lfloor t/V\rfloor$, $\delta_t=(t-t_s)/V$
 \If{$t=t_s$}
  \State Retain at most $N_{\mathrm{slow}}-1$ complete slow blocks and their scores
  \State Form $\operatorname{Read}_{\mathrm{VLM}}(\mathcal{K}_{<t})$ and encode $\mathcal{K}_c$
  \State Append full $\mathcal{K}_c$ to slow memory; update prefill scores
  \State Form $\widehat{\mathcal{K}}_t^{\,\mathrm{act}}$ by Eq.~\ref{eq:memory_views}
  \State Set active consumer $g\gets s$
 \Else
  \State Encode $\mathcal{Z}_t$ and refine $\mathcal{K}_c(t_s)$ to obtain the complete $\widetilde{\mathcal{Z}}_t$
  \State Update the full fast FIFO using Eq.~\ref{eq:fast_fifo}
  \State Form $\widehat{\mathcal{K}}_t^{\,\mathrm{act}}$ by Eq.~\ref{eq:dual_rate_condition}
  \State Set active consumer $g\gets f$
 \EndIf
 \State Generate $A_t$ using fixed $\widehat{\mathcal{K}}_t^{\,\mathrm{act}}$ and current state
 \State Update consumer-$g$ scores using the completed flow solve
 \If{$g=s$}
  \State Clear fast FIFO and fast-denoising scores
 \EndIf
 \State Execute the first $E$ actions of $A_t$
\EndFor
\end{algorithmic}
\end{algorithm}

\subsection{Training Objectives and Optimization}
\label{app:training_details}

\paragraph{Grouped supervision.}
Every dataset frame is eligible as a slow anchor. Each grouped sample supplies one slow branch and two subsequent fast branches in chronological order, each with an $H$-action target. The anchor branch uses the full slow condition, and no refined block from the anchor observation is inserted as a valid fast entry. Later branches use the available uncompressed refined FIFO in chronological order. The slow history is encoded once per grouped sample. Each branch has its own current images, state, target actions, noise, and flow time. Missing historical blocks, unavailable future branches, and action targets beyond the episode boundary are masked.

\paragraph{Flow-matching objective.}
For branch $j$, let $\mathcal{K}_j^{\mathrm{train}}$ denote its complete,
uncompressed observation-KV condition, drawn from the slow cache or
refined FIFO according to the branch. For normalized target actions
$A_j$ and independent standard Gaussian noise $\epsilon_j$, training constructs
\begin{equation}
 \widetilde\tau_j\sim\operatorname{Beta}(1.5,1),\quad
 \tau_j=0.999\widetilde\tau_j+0.001,\quad
 A_j^{\tau_j}=(1-\tau_j)A_j+\tau_j\epsilon_j .
 \label{eq:flow_interpolation}
\end{equation}
The per-component velocity error and grouped loss are
\begin{equation}
 e_{jhd}=v_\theta(A_j^{\tau_j},\tau_j;\mathcal{K}_j^{\mathrm{train}},s_j)_{hd}
 -(\epsilon_j-A_j)_{hd}.
 \label{eq:app_velocity_error}
\end{equation}
\begin{equation}
 \mathcal L_{\mathrm{dual}}
 =\mathbb E\left[
 \frac{\sum_{j,h}m_{jh}\,D^{-1}\sum_{d=1}^{D}e_{jhd}^{2}}
      {\max(1,\sum_{j,h}m_{jh})}\right],
 \qquad D=32 .
 \label{eq:dual_training_loss}
\end{equation}
Here $h$ indexes the $H$ action steps, $d$ indexes the $D$ action
coordinates, $s_j$ is the current state of branch $j$, and $m_{jh}$
masks invalid action steps. Here $j$ is a branch index, not an environment time: $A_j$ is the normalized target chunk beginning at branch time $t_j$, and $s_j=s_{t_j}$ is its current state in the model's normalized, padded representation.
Normalization is performed per anchor
before averaging across the batch. No additional adapter loss is used. The adapter learns through the action objective, and online KV selection is not applied during training.

\paragraph{Initialization and gradient paths.}
% Training starts from the original $\pi_{0.5}$ Base parameters, not a previously fine-tuned causal or DOMINO-Long checkpoint. Full fine-tuning leaves the VLM, action expert, adapter, and state/action projections trainable, and slow KV is not detached. However, the high-rate visual tokens are explicitly stop-gradient outputs of SigLIP. The visual encoder is therefore trained through the slow path, not directly through the high-rate feature path. This distinction is retained even under full fine-tuning.
Training starts from the original $\pi_{0.5}$ Base model. During full fine-tuning, the VLM, action expert, adapter, and state/action projections are trainable, with gradients flowing through the slow-path KV features. In contrast, the high-rate SigLIP~\cite{siglip} features are detached, thus the visual encoder is updated only through the slow path.

\paragraph{Flow integration.}
% Inference uses ten Euler steps from Gaussian noise to the action chunk:
% \begin{equation}
%  X^{(0)}\sim\mathcal N(0,I),\qquad
%  X^{(k+1)}=X^{(k)}-\frac{1}{N_{\mathrm{denoise}}}
%  v_\theta\left(X^{(k)},1-\frac{k}{N_{\mathrm{denoise}}};\widehat{\mathcal{K}}_t^{\,\mathrm{act}},s_t\right).
%  \label{eq:euler_solver}
% \end{equation}
% Here $N_{\mathrm{denoise}}=10$ and $k=0,\ldots,N_{\mathrm{denoise}}-1$; the normalized output is $A_t=X^{(N_{\mathrm{denoise}})}$ before denormalization and conversion to executable actions. The read-view KV remains fixed within the solve. Each call predicts $H$ actions and executes a chunk of size $E=H$ before replanning.

Inference starts from Gaussian noise and uses 10 Euler steps to generate the action chunk:
\begin{equation}
 X^{(0)}\sim\mathcal N(0,I),\qquad
 X^{(k+1)}=X^{(k)}-\frac{1}{N_{\mathrm{denoise}}}
 v_\theta\left(X^{(k)},1-\frac{k}{N_{\mathrm{denoise}}};\widehat{\mathcal{K}}_t^{\,\mathrm{act}},s_t\right),
 \label{eq:euler_solver}
\end{equation}
where $N_{\mathrm{denoise}}=10$. The final output $A_t=X^{(N_{\mathrm{denoise}})}$ is denormalized and converted into executable actions. The read-view KV stays fixed during denoising. Each call predicts and executes an $H$-step action chunk before replanning, i.e., $E=H$.

\subsection{Experiment-Specific Configurations}
\label{app:experiment_configs}

Table~\ref{tab:training_settings} lists the shared optimizer and model settings, with training budgets separated by evaluation setting. DOMINO-Long uses all 500 demonstrations from the ten-task dataset, comprising 227,208 frames, with dataset-specific quantile normalization statistics. A global batch of 40 anchors produces tensors of shape $[40,3,H,32]$, corresponding to 120 action branches before validity masking. Dataset metadata timestamps do not determine the simulation or robot control frequency.

\begin{table}[t]
\centering
\caption{Training and inference settings for DOMINO, DOMINO-Long, and the real-world tasks. Training steps count optimizer updates. Capacities include the current block.}
\label{tab:training_settings}
\begin{tabular*}{\linewidth}{@{\extracolsep{\fill}}ll@{}}
\toprule
Setting & Value\\
\midrule
Initialization / training & $\pi_{0.5}$ Base / full fine-tuning\\
Slow / fast cache capacity & $4/2$ complete blocks\\
Action horizon $H$ / executed chunk size $E$ & $25/25$\\
Slow-update period & $V=3E$\\
Main-result historical visual retention & 0.7 for all active consumers\\
Supervised branches & 3\\
Global batch size & 40 anchors\\
Optimizer & AdamW\\
$(\beta_1,\beta_2)$ / $\epsilon$ & $(0.9,0.95)$ / $10^{-8}$\\
Weight decay / global gradient clipping & $10^{-10}$ / $1.0$\\
Warmup & 1,000 updates\\
Peak / terminal learning rate & $2.5\times10^{-5}$ / $2.5\times10^{-6}$\\
Learning-rate schedule & Warmup followed by cosine decay\\
Parameter EMA / random seed & $0.99$ / 42\\
Compute / parameter precision & BF16 / FP32\\
Flow solver / integration steps & Euler / 10\\
History selection during training & Disabled\\
History selection during inference & Online-H2O, all three consumers\\
\midrule
DOMINO training updates & 75,000\\
DOMINO-Long training updates & 50,000\\
Real-world training updates & 50,000\\
\bottomrule
\end{tabular*}
\end{table}

% The documented DOMINO-Long run uses eight A100 80\,GB GPUs with eight-way FSDP, 20 data-loader workers, prefetch factor two, ordered loading, persistent workers, and no pinned memory. Its cosine schedule spans 50,000 updates including warmup. Checkpoints are saved every 2,000 updates with model parameters, training state, and normalization assets; inference parameters are the EMA weights. These hardware and checkpoint details describe the supplied DOMINO-Long run.
The DOMINO-Long model is trained on eight A100 80 GB GPUs using eight-way FSDP. Training runs for 50,000 updates with a cosine learning-rate schedule, including a warmup stage. For inference, we use the EMA model weights. These settings correspond to the DOMINO-Long configuration used in our experiments.

% TODO: Confirm the DOMINO cosine decay_steps and per-setting hardware from run logs.

\section{Simulation Evaluation Protocols}
\label{app:simulation_protocols}
\label{app:detailed_results}

This appendix describes detailed experimental results for the four simulation benchmarks. Per-task results for DOMINO and RoboTwin are reported below; DOMINO-Long results and LIBERO-Long comparisons are provided in Tab.~\ref{tab:dynamic_domino_results} and Tab.~\ref{tab:robotwin_libero_long}, respectively.

\subsection{DOMINO}
\label{app:domino_protocol}
\label{app:domino_results}
We evaluate DOMINO in the \textbf{L1 + clean} setting. The evaluation covers the 35 tasks listed in Table~\ref{tab:app_domino_all}. We report complete-task success rate (SR) and the benchmark's Manipulation Score (MS), retaining the benchmark scoring conventions. Per-task entries are kept separate from the aggregate comparison in the main text. SR is reported in percent, and MS follows the benchmark definition. Training, memory capacities, and action-chunk settings are specified in Appendix~\ref{app:experiment_configs}.
% TODO: Enter evaluation trial counts, seeds, episode limits, and the exact L1/clean configuration identifier.

\clearpage
\vspace{-4pc}
\begin{table}[H]
\centering
\caption{Per-task DOMINO results under L1 + clean evaluation. Baseline SR (\%) and MS for $\pi_{0.5}$ and PUMA are transcribed from Table 16 of the DOMINO paper; D$^2$-VLA results use a historical visual KV retention ratio of 0.7. All scores are displayed to two decimal places.}
% Source: DOMINO_2603.15620v3.pdf, Table 16, pp. 38--39; L1 + clean setting confirmed by Table 3, p. 11.
% D2-VLA source: compression_r070_all35_task_sr_ms_estimated100_rounded_20260925.csv; SR uses the supplied integer-rounded entries, not evidence of 100 evaluated episodes per task.
\vspace{-.8pc}
\label{tab:app_domino_all}
\small\normalfont
\setlength{\tabcolsep}{4pt}
\renewcommand{\arraystretch}{1.03}
\begin{tabularx}{\linewidth}{@{}l*{6}{>{\centering\arraybackslash}X}@{}}
\toprule
\multirow{2}{*}{\textbf{Task}}
& \multicolumn{2}{c}{$\pi_{0.5}$}
& \multicolumn{2}{c}{\textbf{PUMA}}
& \multicolumn{2}{c}{\textbf{D$^2$-VLA}}\\
\cmidrule(lr){2-3}\cmidrule(lr){4-5}\cmidrule(lr){6-7}
& SR$(\%)\uparrow$ & MS$(\%)\uparrow$ & SR$(\%)\uparrow$ & MS$(\%)\uparrow$ & SR$(\%)\uparrow$ & MS$(\%)\uparrow$ \\
\midrule
Adjust Bottle & 52.00 & 71.45 & 65.00 & 74.04 & 75.00 & 78.18 \\
Beat Block Hammer & 10.00 & 23.22 & 15.00 & 29.09 & 23.00 & 32.29 \\
Click Alarmclock & 14.00 & 18.47 & 4.00 & 8.38 & 5.00 & 8.74 \\
Click Bell & 0.00 & 6.35 & 3.00 & 13.64 & 11.00 & 19.00 \\
Dump Bin Bigbin & 0.00 & 16.64 & 0.00 & 39.05 & 38.00 & 44.32 \\
Grab Roller & 10.00 & 34.32 & 33.00 & 56.48 & 38.00 & 42.05 \\
Handover Block & 1.00 & 33.95 & 17.00 & 64.79 & 13.00 & 46.45 \\
Handover Mic & 5.00 & 29.10 & 35.00 & 54.40 & 46.00 & 59.31 \\
Hanging Mug & 5.00 & 25.59 & 9.00 & 41.74 & 26.00 & 42.34 \\
Move Can Pot & 7.00 & 40.34 & 22.00 & 47.80 & 21.00 & 41.64 \\
Move Pillbottle Pad & 6.00 & 15.93 & 14.00 & 21.82 & 18.00 & 22.81 \\
Move Playingcard Away & 8.00 & 24.84 & 6.00 & 27.90 & 13.00 & 27.14 \\
Move Stapler Pad & 0.00 & 13.38 & 1.00 & 16.24 & 2.00 & 10.35 \\
Place A2B Left & 2.00 & 26.04 & 13.00 & 41.65 & 18.00 & 37.46 \\
Place A2B Right & 1.00 & 23.63 & 8.00 & 34.39 & 18.00 & 35.78 \\
Place Bread Basket & 8.00 & 15.89 & 12.00 & 22.44 & 50.00 & 53.56 \\
Place Bread Skillet & 9.00 & 22.30 & 19.00 & 35.95 & 25.00 & 44.46 \\
Place Can Basket & 6.00 & 29.48 & 14.00 & 45.79 & 14.00 & 34.18 \\
Place Container Plate & 22.00 & 28.46 & 26.00 & 34.45 & 78.00 & 78.53 \\
Place Empty Cup & 2.00 & 12.71 & 7.00 & 17.85 & 24.00 & 30.67 \\
Place Fan & 2.00 & 11.72 & 8.00 & 14.36 & 2.00 & 9.59 \\
Place Mouse Pad & 5.00 & 19.03 & 2.00 & 17.86 & 16.00 & 26.22 \\
Place Object Basket & 11.00 & 30.19 & 13.00 & 41.67 & 41.00 & 50.04 \\
Place Object Scale & 3.00 & 15.37 & 4.00 & 16.05 & 18.00 & 24.80 \\
Place Object Stand & 1.00 & 8.97 & 1.00 & 10.66 & 59.00 & 60.98 \\
Place Phone Stand & 2.00 & 17.98 & 6.00 & 19.02 & 32.00 & 38.57 \\
Place Shoe & 25.00 & 32.84 & 16.00 & 27.23 & 64.00 & 69.63 \\
Press Stapler & 6.00 & 13.27 & 10.00 & 18.40 & 36.00 & 42.75 \\
Put Bottles Dustbin & 17.00 & 33.30 & 23.00 & 36.82 & 14.00 & 20.41 \\
Put Object Cabinet & 11.00 & 46.24 & 34.00 & 61.64 & 41.00 & 66.63 \\
Rotate QRcode & 4.00 & 17.90 & 14.00 & 29.27 & 13.00 & 24.83 \\
Scan Object & 1.00 & 20.48 & 5.00 & 30.58 & 5.00 & 24.21 \\
Shake Bottle & 33.00 & 49.40 & 55.00 & 65.37 & 47.00 & 57.42 \\
Shake Bottle Horizontally & 42.00 & 62.72 & 75.00 & 80.57 & 61.00 & 79.53 \\
Stamp Seal & 6.00 & 22.58 & 13.00 & 26.50 & 21.00 & 36.48 \\
\midrule
% Source discrepancy: Table 16 and Table 3 print pi0.5 mean MS = 26.17; the listed task values average to 26.12, as printed in Table 20. Preserve Table 16 here.
\textbf{Average} & 9.63 & 26.17 & 17.20 & 34.97 & 29.31 & 40.61 \\
\bottomrule
\end{tabularx}
\end{table}

\vspace{-2pc}
\subsection{DOMINO-Long}
\label{appendix:domino_long_dataset}

\begin{figure}[htbp]
    \centering
    \includegraphics[width=0.99\linewidth]{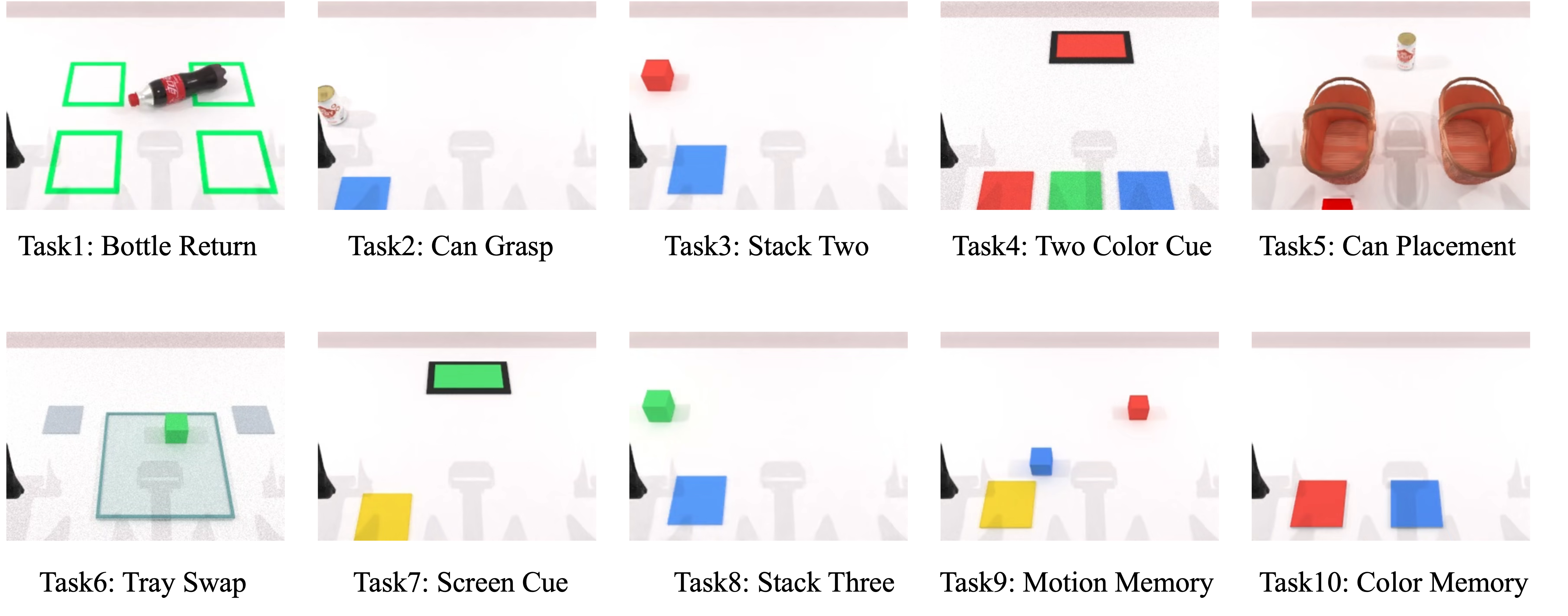}
    \caption{
    \textbf{Overview of DOMINO-Long tasks.}
    Visualization of the ten long-horizon manipulation tasks introduced in DOMINO-Long.
    }
    \label{fig:domino_long}
    \vspace{-2pc}
\end{figure}

\begin{wraptable}{r}{0.65\linewidth}
\vspace{-.2pc}
\centering
% \captionsetup{font=footnotesize,skip=3pt}
\caption{
DOMINO-Long tasks and their memory requirements.
}
\scriptsize
\setlength{\tabcolsep}{2.8pt}
\renewcommand{\arraystretch}{1.10}
\vspace{-.6pc}
\resizebox{\linewidth}{!}{
\begin{tabular}{@{}rll@{}}
\toprule
\textbf{ID} & \textbf{Task} & \textbf{Information retained for later action} \\
\midrule
1  & Bottle Return & Original region of the moving bottle. \\
2  & Can Grasp & Identity of the previously observed can. \\
3  & Stack Two & Presentation order of two blocks. \\
4  & Two Color Cue & Colors specifying the object and destination. \\
5  & Can Placement & Basket indicated by a transient visual cue. \\
6  & Tray Swap & Original source region of the transferred block. \\
7  & Screen Cue & Screen color specifying the target block. \\
8  & Stack Three & Presentation order of three blocks. \\
9  & Motion Memory & Object that previously moved leftward. \\
10 & Color Memory & Association between a color and a region. \\
\bottomrule
\vspace{-1.4pc}
\end{tabular}
}

\label{tab:domino_long_tasks}
\vspace{-0.6em}
\end{wraptable}

DOMINO-Long~\cite{dai2026robommebenchmarkingunderstandingmemory}
comprises 10 simulated tabletop manipulation tasks that combine
visual memory with manipulation of moving objects. The robot must
retain information from earlier observations and use it to determine
later manipulation goals. Such information includes object locations,
identities, motion directions, presentation order, and color--region
associations.

The benchmark uses a simulated Aloha-AgileX dual-arm robot with one
head camera and two wrist cameras. The tasks introduce temporal
dependence in different ways. In cue-based tasks, relevant visual cues
disappear before execution; in sequence-based tasks, objects are
presented earlier and later reappear in different spatial arrangements;
and in manipulation-dependent tasks, previous actions modify the scene
and must be remembered for later decisions.

% The robot must therefore preserve task-relevant history while tracking
% and manipulating moving objects. All ten tasks are illustrated in
% Fig.~\ref{fig:domino_long}, with their main memory requirements
% summarized in Tab.~\ref{tab:domino_long_tasks}.
The robot must preserve relevant history while interacting with moving objects. The 10 tasks and their memory requirements are shown in Fig.~\ref{fig:domino_long} and Tab.~\ref{tab:domino_long_tasks}.

\subsubsection{Task Definitions}
\label{sec:domino_long_definitions}

\paragraph{Task 1: Bottle Return.}
Four visually identical square regions mark possible bottle origins on the
table. A bottle initially occupies one of these regions and subsequently moves
away from it. The robot must intercept and grasp the bottle, orient it upright,
and return it to the region that it originally occupied. The required final
configuration is an upright, released bottle stably contained within that
region. The initial region varies across demonstrations using a balanced
schedule over the four candidates, while the region layout remains fixed.
This task requires retaining an object's original location as its current
position changes, and using that remembered location as the placement goal.

\paragraph{Task 2: Can Grasp.}
A single can first moves across the scene and exits the visible workspace.
After an interval, the same can returns together with two distractor cans.
The robot must recognize the previously observed can, grasp it while it moves,
and place it upright at a designated drop-off location. The target identity
is sampled from a set of six can appearances, and two distinct distractor
identities are selected from the remaining candidates. The lane used during
the initial presentation and the target's lane during the subsequent encounter
are varied separately, with small lane-position perturbations. The task
therefore requires matching object identity across two encounters despite a
change in spatial context and the introduction of distractors.

\paragraph{Task 3: Stack Two.}
A red block and a green block traverse the scene one at a time, establishing
a temporal presentation order. After this presentation, both blocks re-enter
the workspace together in a randomized spatial arrangement. The robot must
construct a two-block stack at a designated stacking area: the first block
observed must form the base, and the second must be placed on top. The
presentation order and the assignment of colors to spatial slots are varied
separately, so the required stacking order cannot be read directly from the
blocks' current left-to-right or lane arrangement. Correct completion requires
the ordered stack to remain stable after release. This task combines memory
for a two-item sequence with moving-object acquisition and precise placement.

\paragraph{Task 4: Two Color Cue.}
A screen presents two color cues in sequence, separated by a blank interval.
The first cue specifies which colored block to manipulate, whereas the second
specifies the destination region. Both cues disappear before manipulation.
The robot subsequently encounters red, green, and blue moving blocks and must
place the block specified by the first cue into the region specified by the
second. The two cue colors are sampled independently and may coincide; the
block-to-lane assignment also varies. The three destination regions retain
their visible colors and fixed color-to-location mapping. The central memory
requirement is to preserve both cue values and their distinct semantic roles,
binding the first to object selection and the second to destination selection.

\paragraph{Task 5: Can Placement.}
Two visually identical baskets occupy the left and right sides of the
workspace. A transient red marker appears in front of one basket to indicate
the intended destination and is then removed. The robot must retain this
destination information while acquiring a moving can, and deposit the can
inside the previously indicated basket. Correct completion requires releasing
the can into the cued basket rather than the alternative basket. The cued
side and the can's motion-side configuration vary across demonstrations,
while the basket and can models remain fixed. This task tests whether a
transient spatial instruction can be maintained through the intervening
tracking and grasping actions.

\paragraph{Task 6: Tray Swap.}
Two stationary source regions are located on opposite sides of the table.
Block $A$ starts in one of these regions, while block $B$ is carried by a
moving tray. The robot must first transfer $A$ onto the tray and then retrieve
$B$ from the tray and place it into the region originally occupied by $A$.
Once $A$ has been removed, its source region is empty; the robot must retain
this source assignment through the intervening transfer. The source side,
the tray's initial direction of travel along the table, and the distinct
colors assigned to the two blocks vary across demonstrations. Completion
requires both parts of the exchange: $A$ is carried by the tray and $B$ is
released at $A$'s original source region.

\paragraph{Task 7: Screen Cue.}
A screen briefly displays red, green, or blue and subsequently turns off.
Three blocks with these respective colors then move through the workspace.
The robot must remember the displayed color, grasp the matching block, and
place it at a fixed designated destination. The cue color and the permutation
of block colors across spatial slots vary across demonstrations using a
balanced sampling schedule. Consequently, the target is specified by the
earlier screen cue rather than by a fixed lane. Unlike Task~4, this task
requires remembering only one color instruction: it selects the object, while
the destination remains unchanged across trials.

\paragraph{Task 8: Stack Three.}
Red, green, and yellow blocks traverse the scene individually in a sampled
order. They subsequently re-enter together with a separately sampled assignment
to spatial slots. The robot must reconstruct the presentation order as a
vertical stack: the first observed block forms the bottom layer, the second
forms the middle layer, and the third forms the top layer. All six temporal
orders and all six spatial permutations are included in the randomization
scheme. Correct completion requires a stable, released tower with the
specified bottom-to-top ordering. Relative to Task~3, this task increases
the number of identities whose order must be retained and extends the
manipulation sequence to three object acquisitions and placements.

\paragraph{Task 9: Motion  Memory.}
A red block and a blue block initially traverse the scene in opposite
directions: one moves from left to right and the other from right to left.
After the demonstration and an intervening interval, the blocks reappear
with a randomized lane assignment. During this retrieval stage, both blocks
move rightward. The robot must select the block that moved leftward during
the earlier demonstration, grasp it, and place it at a fixed destination.
The assignment of colors to initial motion directions, the demonstration
lane assignment, and the target's retrieval lane vary across demonstrations.
The task therefore requires associating an object identity with its previous
motion direction; its direction at the time of grasping does not identify
the target.

\paragraph{Task 10: Color Memory.}
Two spatial regions briefly display red and blue, with one color assigned
to each region. The colored cues then disappear, leaving visually identical
neutral destination regions. A single red or blue query block subsequently
enters the workspace. The robot must grasp this block and place it into the
region that previously displayed the same color. The left--right assignment
of the two region colors, the query-block color, and its entry lane vary
across demonstrations. Correct completion requires releasing the block
within the matching region. This task tests retention of a color-to-location
association: the visible query color must retrieve a destination from an
earlier scene, after the regions themselves no longer reveal their colors.

\subsection{Randomization and Demonstration Collection}
\label{sec:domino_long_collection}

Randomization is defined separately for each task to vary the relationship
between the remembered information and the required action. Depending on
the task, randomized factors include object identity, cue color, source or
destination side, presentation order, motion-direction assignment, and
object-to-lane assignment. Several tasks use balanced schedules over discrete
factor combinations to improve coverage within a finite demonstration set.
The suite does not assume that every scene attribute is randomized:
destination layouts, camera configurations, and other environmental settings
can remain fixed. These variations reduce specific correlations, such as
always encountering a target in the same lane, but do not by themselves
establish the absence of all alternative visual or behavioral cues.

The dataset used for the reported experiments contains 50 demonstrations
for each of the ten tasks, totaling 500 demonstrations. All 500 demonstrations
are used for training, as described in Section~\ref{sec:exp-domino-extension}
and Appendix~\ref{app:experiment_configs}.

\subsection{LIBERO-Long}
\label{app:libero_protocol}
\label{app:libero_results}
LIBERO-Long provides a complementary long-horizon simulation benchmark. We evaluate all ten tasks with the high-rate pathway disabled, using full-KV conditioning during training and online historical selection for the VLM and action expert during inference. Results are reported in Tab.~\ref{tab:robotwin_libero_long}.

\subsection{RoboTwin 2.0}
\label{app:robotwin_protocol}
\label{app:robotwin_results}
We evaluate all 50 RoboTwin~2.0 tasks in the clean setting as a supplementary bimanual simulation benchmark. The high-rate pathway is disabled; training uses full-KV conditioning, and inference uses online historical selection for the VLM and action expert. Tab.~\ref{tab:app_robotwin_all2} reports complete-task SR for each task, with task names corresponding to the environment identifiers.

\begin{table*}[t]
\centering
\caption{
Complete-task success rate (SR, \%) on all 50 RoboTwin~2.0~\cite{robotwin2} tasks.
Task names are formatted from the corresponding environment identifiers.
}
\label{tab:app_robotwin_all2}
\vspace{-0.8pc}
\footnotesize
\setlength{\tabcolsep}{3.5pt}
\renewcommand{\arraystretch}{1.08}

\begin{tabularx}{\textwidth}{
@{}
Y r
@{\hspace{8pt}}
Y r
@{\hspace{8pt}}
Y r
@{}
}
\toprule

\textbf{Task} & \textbf{SR}
&
\textbf{Task} & \textbf{SR}
&
\textbf{Task} & \textbf{SR}
\\

\midrule

Adjust Bottle
& 100\%
&
Open Microwave
& 86\%
&
Place Object Stand
& 90\%
\\

Beat Block Hammer
& 82\%
&
Pick Diverse Bottles
& 48\%
&
Place Phone Stand
& 72\%
\\

Blocks Ranking RGB
& 72\%
&
Pick Dual Bottles
& 62\%
&
Place Shoe
& 82\%
\\

Blocks Ranking Size
& 64\%
&
Place A2B Left
& 70\%
&
Press Stapler
& 86\%
\\

Click Alarmclock
& 100\%
&
Place A2B Right
& 72\%
&
Put Bottles Dustbin
& 68\%
\\

Click Bell
& 100\%
&
Place Bread Basket
& 58\%
&
Put Object Cabinet
& 38\%
\\

Dump Bin Bigbin
& 96\%
&
Place Bread Skillet
& 78\%
&
Rotate QRcode
& 78\%
\\

Grab Roller
& 100\%
&
Place Burger Fries
& 94\%
&
Scan Object
& 34\%
\\

Handover Block
& 62\%
&
Place Can Basket
& 54\%
&
Shake Bottle
& 100\%
\\

Handover Mic
& 96\%
&
Place Cans Plasticbox
& 54\%
&
Shake Bottle Horizontally
& 100\%
\\

Hanging Mug
& 24\%
&
Place Container Plate
& 90\%
&
Stack Blocks Three
& 84\%
\\

Lift Pot
& 96\%
&
Place Dual Shoes
& 62\%
&
Stack Blocks Two
& 96\%
\\

Move Can Pot
& 74\%
&
Place Empty Cup
& 94\%
&
Stack Bowls Three
& 78\%
\\

Move Pillbottle Pad
& 66\%
&
Place Fan
& 72\%
&
Stack Bowls Two
& 94\%
\\

Move Playingcard Away
& 80\%
&
Place Mouse Pad
& 38\%
&
Stamp Seal
& 76\%
\\

Move Stapler Pad
& 28\%
&
Place Object Basket
& 76\%
&
Turn Switch
& 42\%
\\

Open Laptop
& 92\%
&
Place Object Scale
& 58\%
&
&
\\

\midrule

\multicolumn{5}{r}{\textbf{Average SR}}
&
\textbf{74.32\%}
\\

\bottomrule
\end{tabularx}

\end{table*}

\section{Real-World Setup, Tasks, and Evaluation}
\label{sec:real_world_tasks}

We design eight real-world manipulation tasks, organized into four
long-horizon tasks and four long-horizon dynamic tasks.
The first group evaluates the execution of ordered subgoals and
tracking of object assignments across successive manipulations.
The second group is designed to examine how task context is combined
with updated visual observations during multi-stage interaction.

We first describe the real-world platform and the observation--action
interface used for deployment in Appendix~\ref{app:real_platform}. We then introduce the four long-horizon
static tasks in Appendix~\ref{app:real_static} and four long-horizon dynamic tasks in Appendix~\ref{app:real_dynamic}, highlighting the
memory and interaction requirements of each setting. Finally, we define
the complete-task success criterion and report quantitative results
together with representative real-world examples in Appendix~\ref{app:real_results}.

\subsection{Platform and Observation--Action Interface}
\label{app:real_platform}
Experiments use a Unitree G1D robot. The observation interface includes stereo head images and left/right wrist views. For vertical concatenation, each head image is resized to width 224 and height 112 using bilinear interpolation. The left and right images are then stacked vertically, in that order, to form one $224\times224$ head-view input. Together with the two wrist views, this preserves the model's three-view interface rather than introducing a fourth camera-token group.

The Unitree state/action interface contains seven joint dimensions per arm and two gripper dimensions in total. Joint actions are made relative to the branch's current state, whereas gripper actions remain absolute; normalization and padding follow Appendix~\ref{app:model_interfaces}. Real-world training uses 50,000 updates, four slow blocks, and two fast blocks, with action horizon $H$ and executed chunk size $E=H$. Online selection is enabled for VLM prefill and for both slow- and fast-branch denoising at evaluation.
% TODO: Add camera models/mounting, physical control rate, deployment device, teleoperation protocol,
% per-task demonstration counts, and task-to-dataset identifiers. Do not infer these from model periods.

\subsection{Long-Horizon Static Tasks}
\label{app:real_static}

% \begin{figure}[htbp]
%     \centering
    
%     \begin{subfigure}{0.99\linewidth}
%         \centering
%         \includegraphics[width=0.99\linewidth]{img/stack_cup.png}
%         \caption{Ordered Cup Stacking.}
%         \label{fig:stackup_a}
%     \end{subfigure}
    
%     \vspace{0.3em}
    
%     \begin{subfigure}{0.99\linewidth}
%         \centering
%         \includegraphics[width=0.99\linewidth]{img/exchange_object.png}
%         \caption{Cross-Plate Object Swap.}
%         \label{fig:exchange_object}
%     \end{subfigure}
    
%     \vspace{0.3em}
    
%     \begin{subfigure}{0.99\linewidth}
%         \centering
%         \includegraphics[width=0.99\linewidth]{img/pick_in_order.png}
%         \caption{Ordered Object Retrieval.}
%         \label{fig:pick_in_order}
%     \end{subfigure}
    
%     \vspace{0.3em}
    
%     \begin{subfigure}{0.99\linewidth}
%         \centering
%         \includegraphics[width=0.99\linewidth]{img/restore_object.png}
%         \caption{Original-Layout Restoration.}
%         \label{fig:restore_object}
%     \end{subfigure}
    
%     \caption{Long-horizon tasks.}
%     \label{fig:long}
% \end{figure}

\begin{figure}[htbp]
    \centering
    
    \begin{subfigure}{0.8\linewidth}
        \centering
        \includegraphics[width=0.99\linewidth]{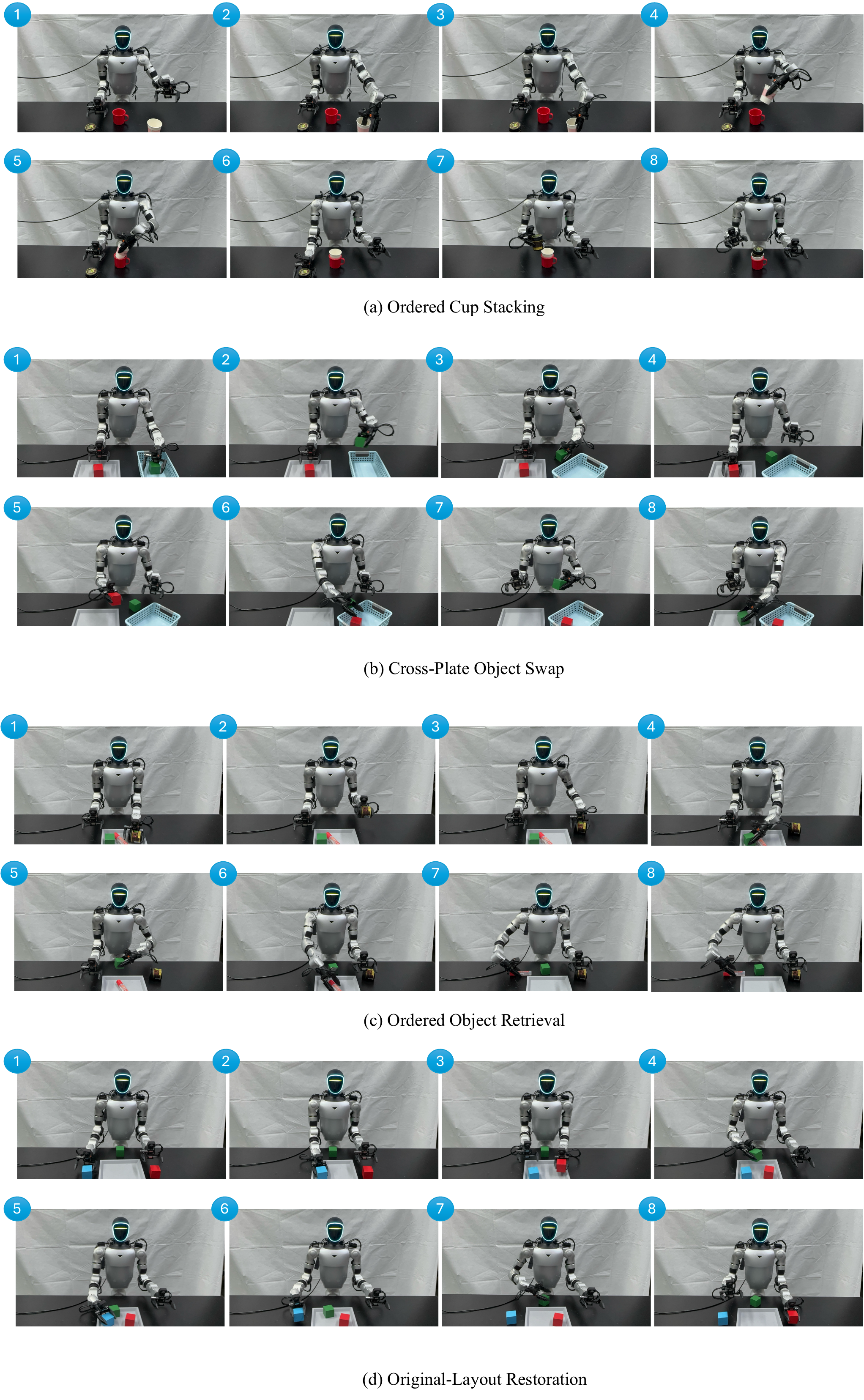}
    \end{subfigure}

    \caption{Long-horizon tasks.}
    \label{fig:long}
\end{figure}

\textbf{(L1) Ordered Cup Stacking.}
The robot stacks a set of cups in a specified order. The task
evaluates adherence to an ordered manipulation sequence and
tracking of progress across successive stacking operations. Real world experiments photos can be seen in Fig.~\ref{fig:long}.

% \begin{figure}[htbp]
%     \centering
%     \includegraphics[width=0.99\linewidth]{img/stack_cup.png}
%     \caption{Ordered Cup Stacking.}
%     \label{fig:stackup}
% \end{figure}

\textbf{(L2) Cross-Plate Object Swap.}
The robot exchanges the objects between two plates so that each
plate ends with the contents initially placed in the other.
The task evaluates object reassignment across a sequence of
pick-and-place operations. Real world experiments photos can be seen in Fig.~\ref{fig:long}.

% \begin{figure}[htbp]
%     \centering
%     \includegraphics[width=0.99\linewidth]{img/exchange_object.png}
%     \caption{Cross-Plate Object Swap.}
%     \label{fig:exchange_object}
% \end{figure}

\textbf{(L3) Ordered Object Retrieval.}
The robot removes objects from a plate in a specified sequence.
The task evaluates whether successive object selections follow
the prescribed order throughout execution. Real world experiments photos can be seen in Fig.~\ref{fig:long}.

% \begin{figure}[htbp]
%     \centering
%     \includegraphics[width=0.99\linewidth]{img/pick_in_order.png}
%     \caption{Ordered Object Retrieval.}
%     \label{fig:pick_in_order}
% \end{figure}

\textbf{(L4) Original-Layout Restoration.}
The robot first places differently colored blocks into a plate,
then retrieves them and returns each block to its original
location. The task evaluates retention of object--location
associations across an intervening rearrangement phase. Real world experiments photos can be seen in Fig.~\ref{fig:long}.

% \begin{figure}[htbp]
%     \centering
%     \includegraphics[width=0.99\linewidth]{img/restore_object.png}
%     \caption{Original-Layout Restoration.}
%     \label{fig:restore_object}
% \end{figure}

\subsection{Long-Horizon Dynamic Tasks}
\label{app:real_dynamic}

\begin{figure}[htbp]
    \centering
    
    \begin{subfigure}{0.8\linewidth}
        \centering
        \includegraphics[width=0.99\linewidth]{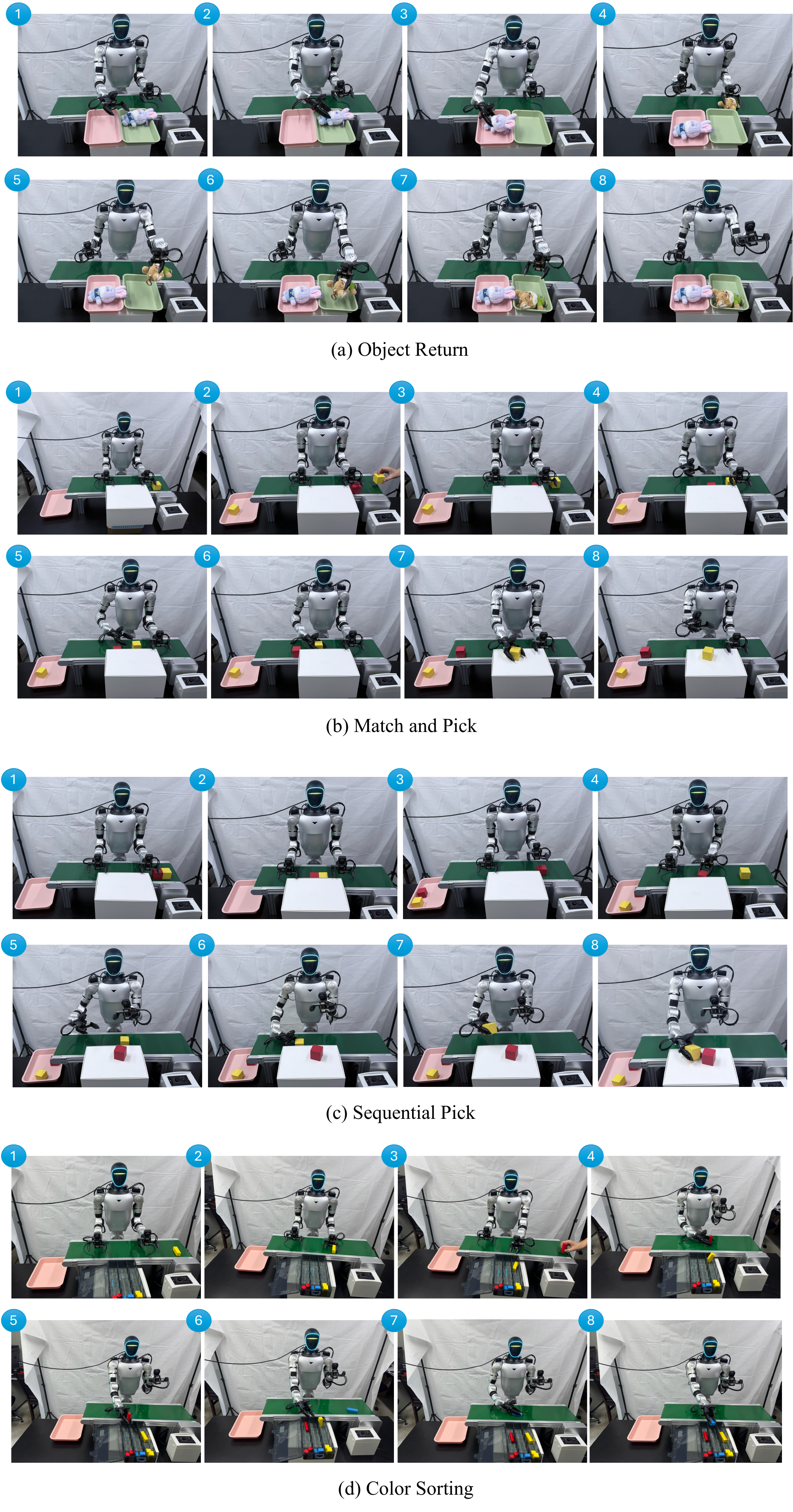}
    \end{subfigure}

    \caption{Long-horizon dynamic tasks.}
    \label{fig:longdynamic}
\end{figure}

\textbf{(D1) Object Return.}
The robot first transfers a toy from one plate to another.
It then grasps a new toy arriving on a conveyor belt and places
it at the original toy's previous location in the source plate.
The task combines an earlier spatial reference with manipulation
of a newly arriving object. Real world experiments photos can be seen in Fig.~\ref{fig:longdynamic}.

% \begin{figure}[htbp]
%     \centering
%     \includegraphics[width=0.99\linewidth]{img/exchange_toy.png}
%     \caption{Object Return.}
%     \label{fig:exchange_toy}
% \end{figure}

\textbf{(D2) Match and Pick.}
The robot first observes a colored reference block passing along
the conveyor. When two differently colored blocks subsequently
arrive, it must grasp the block matching the reference color.
The task couples retention of an earlier color cue with timely
selection of a moving target. Real world experiments photos can be seen in Fig.~\ref{fig:longdynamic}.

% \begin{figure}[htbp]
%     \centering
%     \includegraphics[width=0.99\linewidth]{img/pick_remember_object.png}
%     \caption{Match and Pick.}
%     \label{fig:pick_remember}
% \end{figure}

\textbf{(D3) Sequential Pick.}
The robot first observes two differently colored blocks passing
along the conveyor in sequence. When two blocks of the same colors
subsequently arrive, it must grasp them in the previously observed
color order. The task combines temporal-order recall with
sequential grasping of moving objects. Real world experiments photos can be seen in Fig.~\ref{fig:longdynamic}.

% \begin{figure}[htbp]
%     \centering
%     \includegraphics[width=0.99\linewidth]{img/pick_sequential.png}
%     \caption{Sequential Pick.}
%     \label{fig:pick_sequential}
% \end{figure}

\textbf{(D4) Color Sorting.}
The robot first observes color cues indicating a color-to-compartment
mapping. The cues subsequently disappear or become occluded, requiring
the robot to retain this mapping. Objects move with a conveyor belt;
the robot must grasp them and place each into the compartment associated
with its color. The conveyor provides the scene dynamics, while other
stationary objects have no programmed autonomous changes. The task tests
memory for earlier color--location associations together with timely
manipulation of moving objects. Examples are shown in Fig.~\ref{fig:longdynamic}.

% \begin{figure}[htbp]
%     \centering
%     \includegraphics[width=0.99\linewidth]{img/sort_color.png}
%     \caption{Color Sorting.}
%     \label{fig:sort_color}
% \end{figure}

\subsection{Evaluation Protocol and Additional Examples}
\label{app:real_results}
% The primary metric is complete-task SR. A successful trial must complete the instructed manipulation sequence, including its required object identities, order, and final arrangement where applicable. Intermediate grasp success or correct cue recognition alone does not count as complete success. Fig.~\ref{fig:Real_world_experiment} presents the dynamic-task examples and comparisons. Tab.~\ref{tab:app_real_all} reports results for the four static tasks, with supplementary task demonstrations in Appendices~\ref{app:real_static} and~\ref{app:real_dynamic}.

% \begin{table}[H]
% \centering
% \caption{Complete-task SR (\%) on the four long-horizon static real-world tasks. Dynamic-task demonstrations and results are presented in Fig.~\ref{fig:Real_world_experiment}.}
% \label{tab:app_real_all}
% \small
% \begin{tabular*}{\linewidth}{@{\extracolsep{\fill}}lccc@{}}
% \toprule
% Task & $\pi_{0.5}$ & MemoryVLA & D$^2$-VLA\\
% \midrule
% \multicolumn{4}{l}{Long-horizon static tasks}\\
% L1: Ordered Cup Stacking & 70 &60 &80 \\
% L2: Cross-Plate Object Swap &50 &40 & 75\\
% L3: Ordered Object Retrieval &45 & 35&65 \\
% L4: Original-Layout Restoration & 40& 25& 65\\
% Static average &51.25 &40 & 71.25\\

% \bottomrule
% \end{tabular*}
% \end{table}

\begin{table}[H]
\centering

% ==================== Left: Evaluation Protocol ====================
\begin{minipage}[t]{0.46\linewidth}
\vspace{0pt}
\small

\textbf{Evaluation Protocol.}
We use complete-task success rate (SR) as the primary metric for real-world evaluation. A trial is considered successful only when the full manipulation sequence is completed, including the required object identities, action order, and final configuration when applicable. Intermediate success, such as a correct grasp or cue recognition, does not count as task completion.

Fig.~\ref{fig:Real_world_experiment} presents representative dynamic-task executions and comparisons. Additional demonstrations of the static and dynamic tasks are provided in Appendices~\ref{app:real_static} and~\ref{app:real_dynamic}, respectively.

\end{minipage}
\hfill
% ==================== Right: Static-task Results ====================
\begin{minipage}[t]{0.52\linewidth}
\vspace{0pt}
\centering

\caption{
Complete-task SR (\%) on the four long-horizon static real-world tasks. M-VLA and Ours denotes Memory-VLA and $\text{D}^2-VLA$, respectively.
}
\label{tab:app_real_all}

\footnotesize
\setlength{\tabcolsep}{2pt}
\renewcommand{\arraystretch}{1.08}
\vspace{-0.7pc}
\begin{tabularx}{\linewidth}{@{}>{\raggedright\arraybackslash}Xccc@{}}
\toprule
\textbf{Task}
& $\boldsymbol{\pi_{0.5}}$
& \textbf{M-VLA}
& \textbf{Ours} \\
\midrule

L1: Ordered Cup Stacking
& 70 & 60 & \textbf{80} \\

L2: Cross-Plate Object Swap
& 50 & 40 & \textbf{75} \\

L3: Ordered Object Retrieval
& 45 & 35 & \textbf{65} \\

L4: Original-Layout Restoration
& 40 & 25 & \textbf{65} \\

\midrule
\textbf{Average}
& 51.25 & 40 & \textbf{71.25} \\

\bottomrule
\end{tabularx}

\end{minipage}

\end{table}

\section{Memory Analysis and Ablation Studies}
\label{app:memory_analysis}

We provide detailed analysis of the memory mechanisms in
D$^2$-VLA and examine their effects on both task performance and inference efficiency in this section. We first visualize how the VLM and action expert attend to
historical KV states, highlighting their distinct patterns of historical
context usage. We then study the accuracy--memory--latency trade-off of
historical KV caching by comparing single-observation inference, uncompressed
full-KV reads, and our selective Dual-Memory design. Next, we analyze the
effect of historical KV retention ratios to characterize how aggressively
the cached context can be compressed without sacrificing task performance.
Finally, we ablate the update frequency of the high-rate pathway to evaluate
how intermediate visual updates between VLM refreshes contribute to dynamic
long-horizon control.

\subsection{Historical KV Attention Analysis}
\label{app:attention_analysis}
\begin{figure}[t]
    \centering

    % First row: LIBERO
    \begin{subfigure}[t]{0.48\linewidth}
        \centering
        \includegraphics[width=\linewidth]{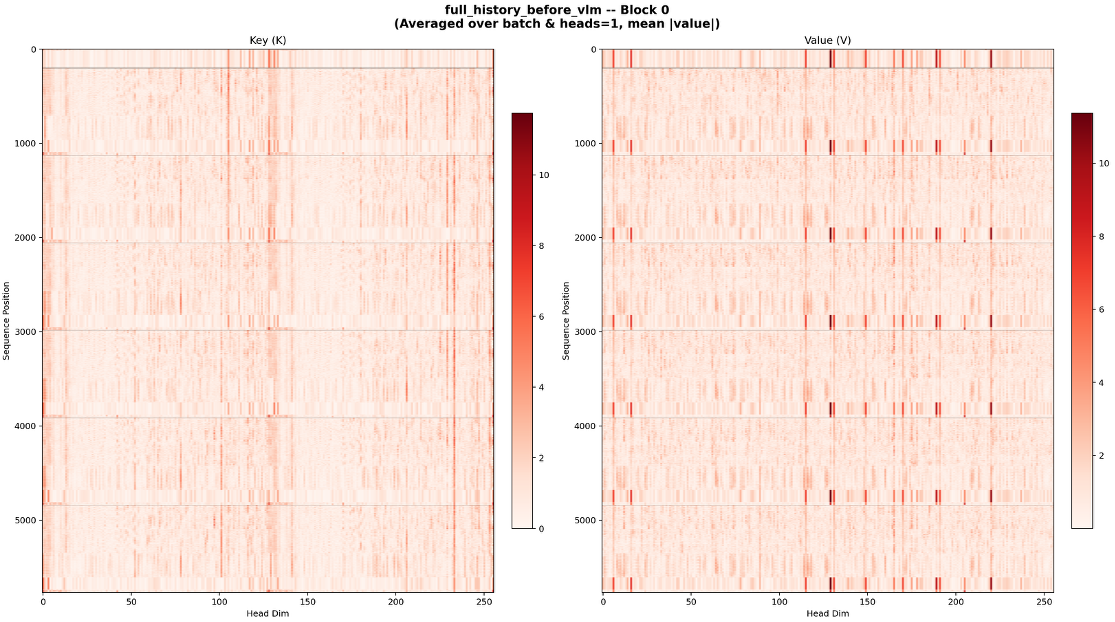}
        \caption{LIBERO, chunk 5.}
        \label{fig:kv1}
    \end{subfigure}
    \hfill
    \begin{subfigure}[t]{0.48\linewidth}
        \centering
        \includegraphics[width=\linewidth]{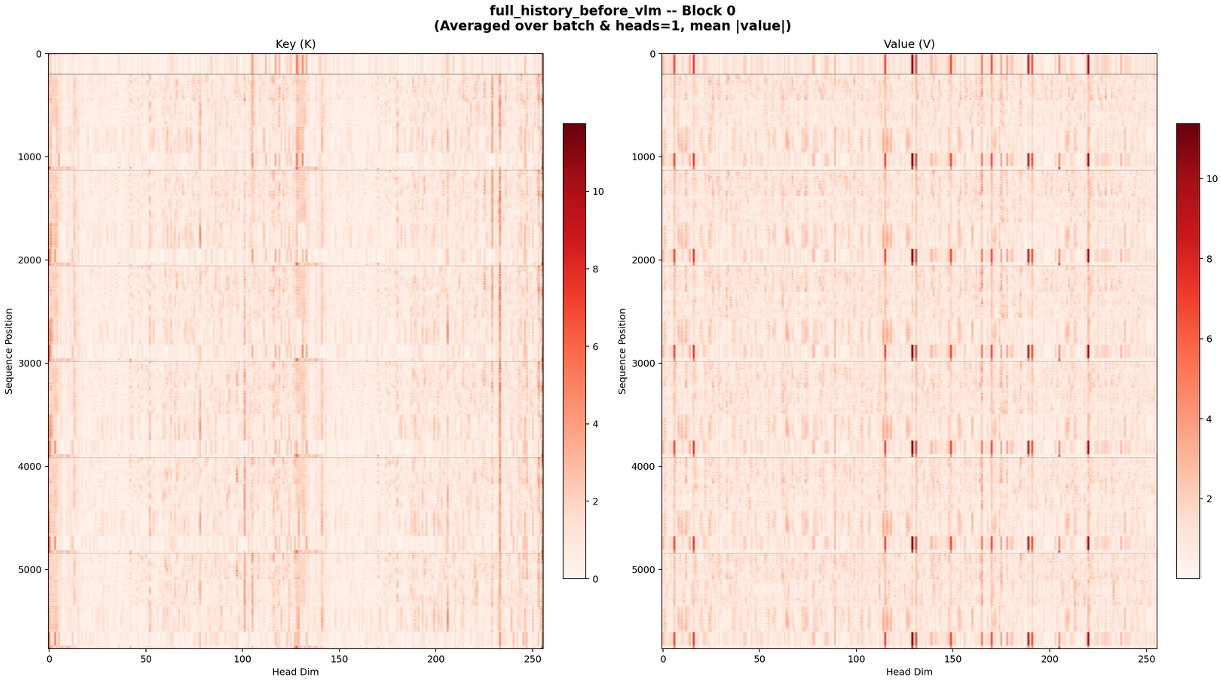}
        \caption{LIBERO, chunk 6.}
        \label{fig:kv2}
    \end{subfigure}

    \par\medskip

    % Second row: RoboTwin
    \begin{subfigure}[t]{0.48\linewidth}
        \centering
        \includegraphics[width=\linewidth]{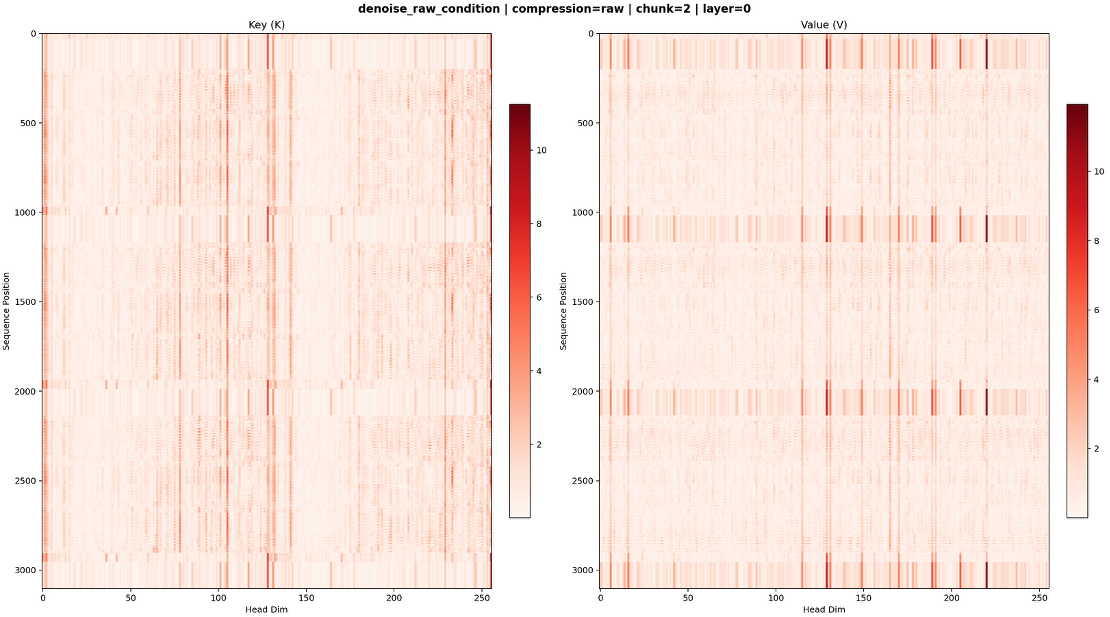}
        \caption{RoboTwin, chunk 2.}
        \label{fig:kv3}
    \end{subfigure}
    \hfill
    \begin{subfigure}[t]{0.48\linewidth}
        \centering
        \includegraphics[width=\linewidth]{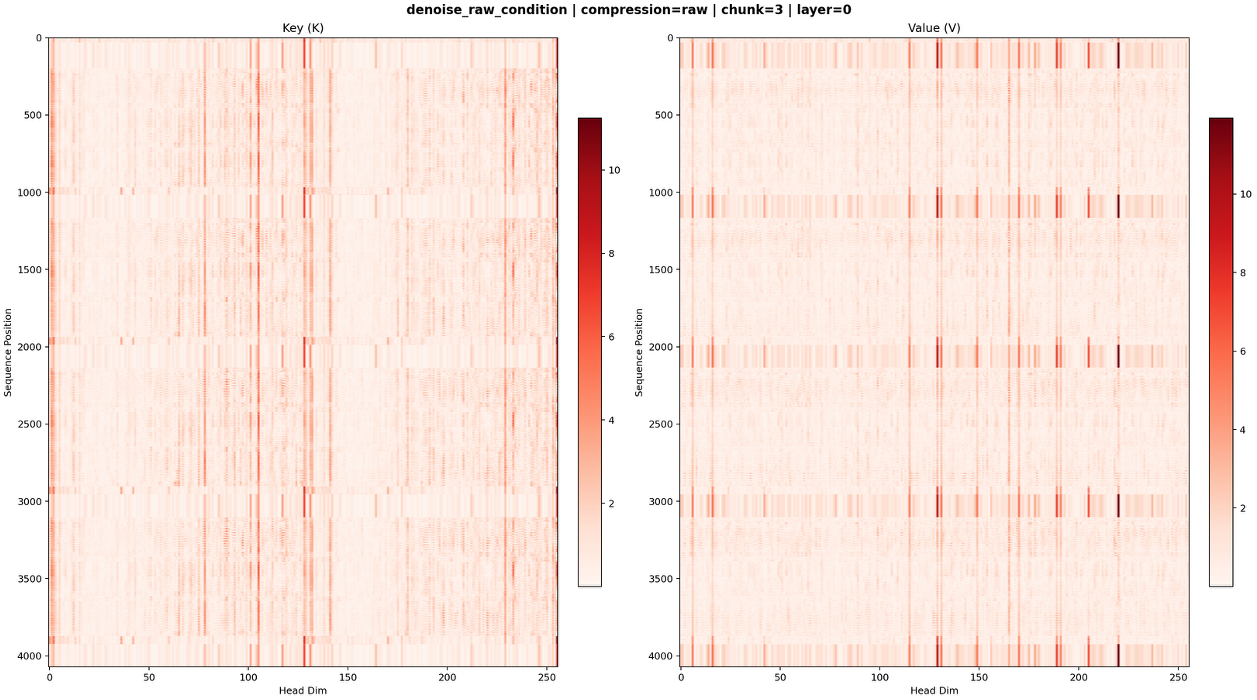}
        \caption{RoboTwin, chunk 3.}
        \label{fig:kv4}
    \end{subfigure}

    \caption{
        \textbf{Historical KV-cache activation distributions on
        LIBERO and RoboTwin.}
        In each subfigure, the left and right heatmaps show the
        magnitudes of cached keys (K) and values (V), respectively.
        The distributions are non-uniform, with stronger activations
        concentrated in a subset of tokens and feature dimensions.
        This concentration motivates selective KV-cache compression
        that prioritizes salient historical entries, suggesting the
        potential to reduce memory and computation with limited
        performance loss.
    }
    \label{fig:kv_analysis}
\end{figure}

Fig.~\ref{fig:kv_analysis} shows the existing visualizations of historical KV attention for LIBERO~\cite{dai2026robommebenchmarkingunderstandingmemory} and RoboTwin~\cite{robotwin2}. They separate the VLM and action-expert read patterns at two observation chunks. These plots are diagnostic visualizations, not an offline prior used by the online selector. These examples supplement the qualitative attention-asymmetry and token-concentration analysis in Section~\ref{sec:dual_memory_selection}. They do not establish that the expert always selects more recent frames: the implemented selector ranks tokens within each retained historical block, without a separate recent-frame-only rule.
% TODO: Recover the plotting records for these existing figures. Do not equate them with the
% ordinary DOMINO L123 72k attention run or the DOMINO-Pro 50k training run without provenance.
% \begin{figure}[t]
%     \centering
%     % Left panel
%     \begin{subfigure}[t]{0.48\linewidth}
%         \centering
%         \vspace{0pt}
%         \includegraphics[width=\linewidth]{img/kv1.png}
%         \caption{LIBERO chunk5 attention distribution.}
%         \label{fig:kv1}
%     \end{subfigure}
%     \hfill
%     % Right panel
%     \begin{subfigure}[t]{0.48\linewidth}
%         \centering
%         \vspace{0pt}
%         \includegraphics[width=\linewidth]{img/kv2.png}
%         \caption{LIBERO chunk6 attention distribution.}
%         \label{fig:kv2}
%     \end{subfigure}

%     \caption{
%         Attention analysis of the VLM and diffusion action expert
%         over historical KV caches.
%     }
%     \label{fig:kv_analysis}
% \end{figure}

% \begin{figure}[t]
%     \centering
%     % Left panel
%     \begin{subfigure}[t]{0.48\linewidth}
%         \centering
%         \vspace{0pt}
%         \includegraphics[width=\linewidth]{img/kv3.png}
%         \caption{RoboTwin chunk2 attention distribution.}
%         \label{fig:kv3}
%     \end{subfigure}
%     \hfill
%     % Right panel
%     \begin{subfigure}[t]{0.48\linewidth}
%         \centering
%         \vspace{0pt}
%         \includegraphics[width=\linewidth]{img/kv4.png}
%         \caption{RoboTwin chunk3 attention distribution.}
%         \label{fig:kv4}
%     \end{subfigure}

%     \caption{
%         Attention analysis of the VLM and diffusion action expert
%         over historical KV caches.
%     }
%     \label{fig:kv_analysis1}
% \end{figure}

\subsection{Accuracy--Memory--Latency Trade-off of KV Caching}
\label{app:kv_cache_tradeoff}

\begin{figure}[t]
\centering

\caption{
\textbf{Accuracy--efficiency and historical KV-retention analysis.}
(a) Comparison with the single-observation baseline and Full KV Cache. LBO-L, RBT2.0, and DMO denote the LIBERO-Long~\cite{dai2026robommebenchmarkingunderstandingmemory}, RoboTwin~2.0~\cite{robotwin2}, and DOMINO~\cite{dai2026robommebenchmarkingunderstandingmemory} benchmarks, respectively. SR, Lat., and Mem. refer to average success rate, inference latency, and peak GPU memory usage; (b) DOMINO success rate under different historical visual KV-retention ratios.}
\label{fig:kv_tradeoff}
\vspace{-.8pc}
\vspace{3pt}

% Left: table
\begin{minipage}[b]{0.43\linewidth}
\centering

\small
\setlength{\tabcolsep}{2.2pt}
\renewcommand{\arraystretch}{1.2}

\begin{tabular}{@{}llccc@{}}
\toprule
\textbf{Data} & \textbf{Method}
& \textbf{SR}$\uparrow$
& \textbf{Lat.}$\downarrow$
& \textbf{Mem.}$\downarrow$ \\
\midrule

\multirow{3}{*}{LBO-L}
& $\pi_{0.5}$ & 92.4 & 141.03 & 9157 \\
& Full KV & 94.5 & 184.69 & 10738 \\
& \textbf{Ours} & \textbf{97.5} & \textbf{174.92} & \textbf{9654} \\
\midrule

\multirow{3}{*}{RBT2.0}
& $\pi_{0.5}$ & 57.0 & 60.10 & 20854 \\
& Full KV & 73.12 & 71.60 & 21973 \\
& \textbf{Ours} & \textbf{74.3} & \textbf{70.10} & \textbf{21068} \\
\midrule

\multirow{3}{*}{DMO}
& $\pi_{0.5}$ & 9.6 & 89.62 & 20962 \\
& Full KV & 28.5 & 93.83 & 22773 \\
& \textbf{Ours} & \textbf{29.3} & \textbf{91.43} & \textbf{21309} \\
\bottomrule
\end{tabular}

\vspace{3pt}
{\footnotesize (a) Accuracy--efficiency comparison.}
\label{tab:kv_cache_latency}
\end{minipage}%
\hspace{0.01\linewidth}%
% Right: figure
\begin{minipage}[b]{0.55\linewidth}
\centering

\includegraphics[width=\linewidth]{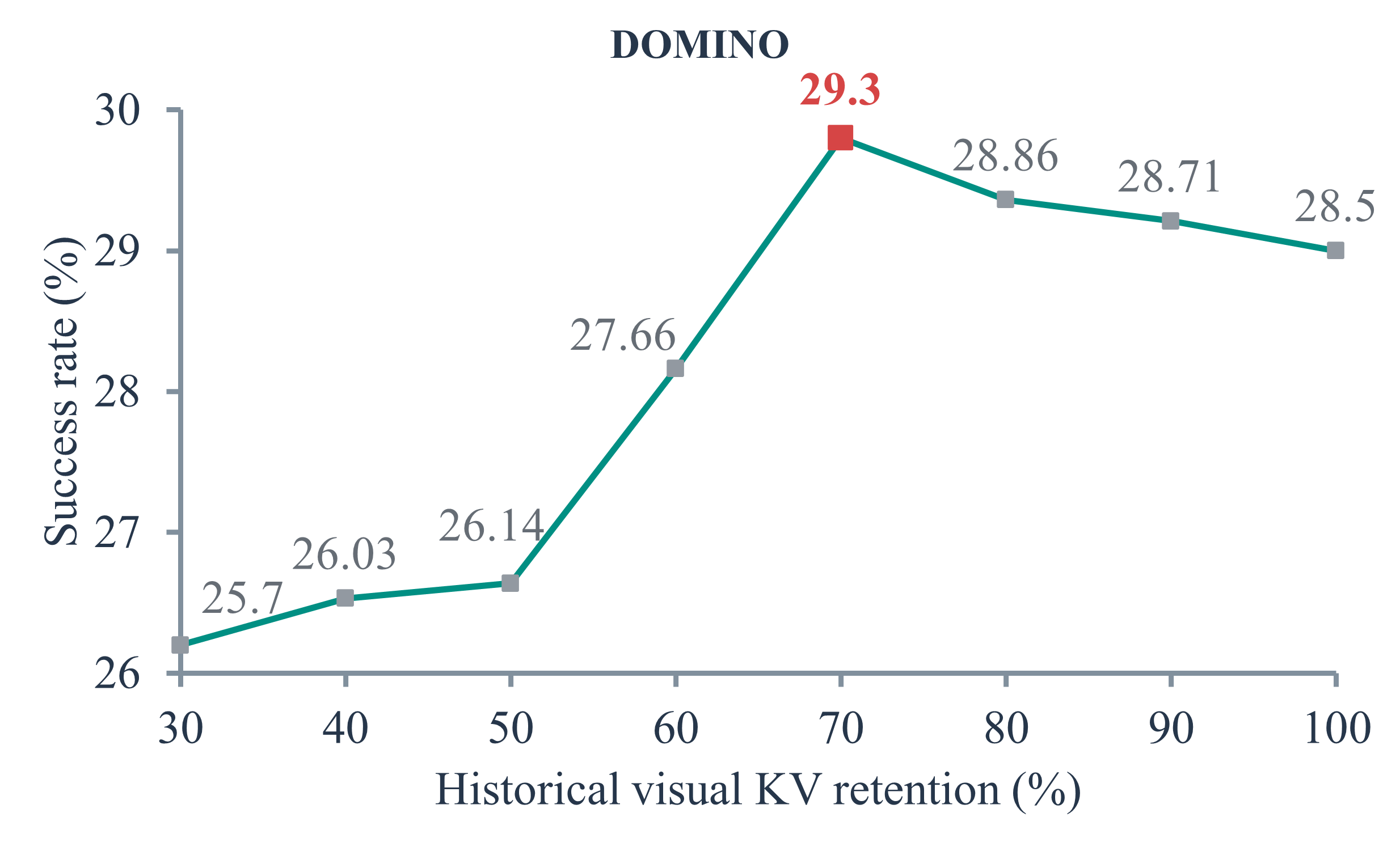}

\vspace{3pt}
{\footnotesize (b) Historical KV-retention analysis.}
\label{fig:app_kv_retention}
\end{minipage}
\vspace{-1.5pc}
\end{figure}

Tab.~\ref{tab:kv_cache_latency} examines the efficiency--performance
trade-off of historical KV memory. \textit{Full KV Cache} denotes
uncompressed historical KV reads for all active consumers in each
benchmark. On LIBERO-Long and RoboTwin 2.0, the high-rate pathway is
disabled, and Full KV Cache uses complete historical KV reads for
VLM prefill and action-expert denoising. On DOMINO, the dual-frequency
pathway remains enabled: VLM prefill and slow-branch denoising read
the complete retained slow-memory history, while fast-branch denoising
reads the complete retained fast-memory history. Thus, the DOMINO
Full KV Cache baseline includes both adapter-based visual updates
and fast memory. D$^2$-VLA applies consumer-specific selection to
the corresponding historical reads, while preserving instruction
tokens and complete current blocks. Both historical-memory variants
retain complete KV blocks in persistent storage; selection changes
only the temporary read views.

Compared with the single-observation $\pi_{0.5}$ baseline, these
benchmark-specific D$^2$-VLA configurations achieve substantially
higher success rates with additional inference cost. On LIBERO-Long,
the success rate increases from 92.4\% to 97.5\%, while latency
increases from 141.03\,ms to 174.92\,ms and peak memory from
9,157\,MiB to 9,654\,MiB. On RoboTwin 2.0 and DOMINO, D$^2$-VLA
improves success by 17.3 and 19.7 percentage points, respectively,
while increasing peak memory by only about 1--2\% over the
single-observation baseline. These results show that historical
conditioning, together with intermediate visual updates on DOMINO,
provides substantial performance gains with comparatively limited
increases in peak memory usage.

Relative to Full KV Cache, D$^2$-VLA achieves higher success rates
on all three benchmarks while reducing both latency and peak memory
usage. On LIBERO-Long, latency decreases from 184.69\,ms to
174.92\,ms and peak memory from 10,738\,MiB to 9,654\,MiB,
corresponding to approximately 5.3\% lower latency and 10.1\% lower
memory, while success increases from 94.5\% to 97.5\%. Similar
trends hold on RoboTwin 2.0 and DOMINO, where peak memory decreases
by approximately 4.1\% and 6.4\%, respectively, with lower inference
latency in both cases. In particular, the DOMINO comparison evaluates
selective versus uncompressed historical reads with the high-rate
pathway enabled in both variants. These results support
consumer-specific historical KV selection over reading all tokens
in the retained blocks.

The comparison also suggests a benefit of selective historical KV
reads beyond computational efficiency. Full KV Cache exposes each
consumer to all tokens in its retained historical blocks, including
potentially irrelevant or redundant context. In contrast, Dual-Memory
constructs separate historical read views using consumer-specific
online attention statistics: VLM prefill and slow denoising select
from slow memory, while fast denoising selects from fast memory
when the high-rate pathway is enabled. The higher success rates
suggest that selective reads may help filter redundant context,
although this comparison alone does not isolate that mechanism.
Overall, D$^2$-VLA provides a favorable balance between task
performance and inference cost through selective historical reads,
without discarding tokens from the retained persistent KV blocks.

\subsection{Consumer-Specific Selection Analysis}
\label{app:selection_analysis}
We evaluate historical visual-token retention ratios of 1.0 (uncompressed), 0.9, 0.8, 0.7, 0.6, 0.5, 0.4, and 0.3. Within each experiment, VLM prefill, slow denoising, and fast denoising use the same ratio while maintaining independent attention statistics. The score-retention coefficient is held fixed at $\rho=0.2$ across this sweep. The main results use a ratio of 0.7. Fig.~\ref{fig:app_kv_retention}~(b) shows DOMINO success rates across historical visual-token retention ratios. Instruction tokens and the current KV block remain uncompressed, so these ratios do not describe the entire denoising prefix. Equal retention ratios do not imply shared scores or identical selected indices. In the current implementation, an uncompressed read can be obtained either with unit retention ratios and no explicit token budgets, or by disabling selection; latency comparisons state that online scoring remains enabled.

% \begin{figure}[t]
% \centering
% \includegraphics[width=0.88\linewidth]{img/fig10.png}
% \caption{Effect of historical visual KV retention on DOMINO success rate. VLM prefill, slow denoising, and fast denoising use the same retention ratio with independent scores and fixed $\rho=0.2$. Instruction tokens, the current denoising block, and complete stored KV are preserved. The main setting retains 70\% of historical visual tokens.}
% \label{fig:app_kv_retention}
% \end{figure}

% \subsection{Inference Cost of Historical KV Selection}
% \label{app:component_ablations}
% Table~\ref{tab:app_component_ablation} compares the inference cost of D$^2$-VLA with uncompressed historical reads. Historical KV selection reduces peak GPU memory usage from 22,773 to 21,309\,MiB (6.4\%) and latency from 93.83 to 91.43\,ms (2.6\%). These results show lower inference memory usage and a modest latency reduction. Selection compacts the historical KV read views rather than the complete blocks retained in persistent storage.

% \begin{table}[t]
% \centering
% \caption{Inference cost with and without historical KV selection. Peak memory denotes peak GPU memory usage during inference. Lower values are better for both metrics.}
% \label{tab:app_component_ablation}
% \small
% \begin{tabular*}{\linewidth}{@{\extracolsep{\fill}}lcc@{}}
% \toprule
% Configuration & Peak memory (MiB) $\downarrow$ & Latency (ms) $\downarrow$\\
% \midrule
% D$^2$-VLA & 21309 & 91.43\\
% Uncompressed historical reads & 22773 & 93.83\\
% \bottomrule
% \end{tabular*}
% \end{table}

\subsection{Update-Frequency Ablation}
\label{app:memory_sensitivity}

\begin{table}[H]
\centering
\caption{Dual-frequency ablation on DOMINO. The enabled settings use $E=25$ and vary the slow-update period $V$.}
\label{tab:ablation_dual_frequency}
\begin{tabular}{lcc}
\toprule
Setting & Period ratio $V/E$ & SR (\%)\\
\midrule
Without dual-frequency mechanism & -- & 16.1\\
$V:E=50:25$ & 2 & 27.6\\
$V:E=75:25$ & 3 & \textbf{29.3}\\
\bottomrule
\end{tabular}
\end{table}

\vspace{-1pc}

Tab.~\ref{tab:ablation_dual_frequency} compares slow-only inference with the dual-frequency mechanism disabled (``--'') against $(V,E)=(50,25)$ and $(75,25)$. The enabled variants fix $H=E=25$, yielding period ratios of 2 and 3 and one or two intermediate adapter calls per slow cycle, respectively. They achieve SRs of 27.6\% and 29.3\%, compared with 16.1\% without the high-rate pathway, supporting the benefit of incorporating fresh observations between VLM updates. The two enabled settings differ in the slow-refresh interval and the number of intermediate updates, not in the absolute action-replanning rate. Our main configuration uses $V=3E$ and $H=E=25$.
% TODO: Add capacity/action-chunking experiments after the evaluated variants are finalized.

\end{document}